\documentclass{article}
\usepackage{ijcai22}

\usepackage{times}
\usepackage{soul}
\usepackage{url}
\usepackage[hidelinks]{hyperref}
\usepackage[utf8]{inputenc}
\usepackage[small]{caption}
\usepackage{graphicx}
\usepackage{amsmath}
\usepackage{amsthm}
\usepackage{booktabs}

\newtheorem{theorem}{Theorem}

\usepackage{caption}
\usepackage{subcaption}
\usepackage{booktabs}
\usepackage{multirow}
\usepackage[ruled,vlined]{algorithm2e}
\SetKwInput{KwInput}{Input}
\usepackage{enumitem}
\usepackage{hyperref}

\usepackage{xcolor}

\newcommand{\rulesep}{\unskip\ \vrule\ }

\usepackage{amsmath,amsthm}
\theoremstyle{definition}
\newtheorem{definition}{Definition}%[section]
\title{fAux: Testing Individual Fairness via Gradient Alignment}

\author{
Giuseppe Castiglione$^1$\and
Ga Wu$^1$\and
Christopher Srinivasa$^1$\And
Simon Prince$^1$\\
\affiliations
$^1$Borealis AI
\emails
\{giuseppe.castiglione, ga.wu, christopher.srinivasa, simon.prince\}@borealisai.com
}

\begin{document}

\maketitle

\begin{abstract}
Machine learning models are vulnerable to biases that result in unfair treatment of individuals from different populations.  Recent work that aims to test a model's fairness at the individual level either relies on domain knowledge to choose metrics, or on input transformations that risk generating out-of-domain samples.  We describe a new approach for testing individual fairness that does not have either requirement.  We propose a novel criterion for evaluating individual fairness and develop a practical testing method based on this criterion which we call \textit{fAux} (pronounced fox).  This is based on comparing the derivatives of the predictions of the model to be tested with those of an {\em auxiliary model}, which predicts the protected variable from the observed data. We show that the proposed method effectively identifies discrimination on both synthetic and real-world datasets, and has quantitative and qualitative advantages over contemporary methods.
\end{abstract}

\section{Introduction}

%Removing \cite{liu2019fair}, as it's older, and just one specific alg that uses RL which we don't directly compare against.
%Ditto for lahoti2019ifair  The references we keep are (1) adversarial representations, and (2) involve similarity metrics, which reviewers are more likely to ask us to compare against.
%Zemel is an influential paper, but it's also an old one.  I have to remove one more paper, I'm going to choose this one.  It's also just...such a long citation...

%Also removing:
%\cite{rajkomar2018ensuring,berk2018fairness,yucer2020exploring}

\noindent  Unfair treatment of different populations by machine learning models can result in undesired social impact~\cite{berk2018fairness,yucer2020exploring}. There are three main research challenges associated with this problem.  The first is to identify the source of the bias and understand how this influences the models \cite{mehrabi2019survey,sun2020evolution}.  The second is to modify the training strategy to prevent unfair predictions  \cite{yurochkin2020training,ruoss2020learning,yurochkin2021sensei}.  The final challenge, which is addressed in this paper, is to test the fairness of existing ML models.  

%Omitted \cite{verma2018fairness}
%This paper addresses individual fairness, but even here there are multiple, potentially conflicting criteria~\cite{verma2018fairness}

To test that a model is fair, we must first agree on what is meant by `fairness'.  For all definitions, fairness is defined with respect to {\em protected variables} such as race, gender, or age.  However, the literature distinguishes between {\em group fairness} (equivalent aggregate treatment of different protected groups) and {\em individual fairness} (equivalent treatment of similar individuals regardless of their protected group). This paper addresses individual fairness, but even here there are multiple, potentially conflicting criteria.   For example, a common definition~(see \cite{gajane2017formalizing}) is  {\em fairness through unawareness}~(FTU) in which the model should behave as though the protected variable is not present. Conversely, \cite{dwork2012fairness} proposed {\em fairness through awareness}~(FTA), which requires that similar individuals have similar prediction outcomes. \cite{CounterfactualFairness} emphasized {\em counterfactual fairness}~(CFF). This takes a data example and synthesizes a counterfactual example in which the protected variable and its descendants are changed. It requires that the original predictions and those for the counterfactual should be similar. 
%\Simon{Are you happy with this definition of a counterfactual example?}

\begin{figure}
    \centering
    \includegraphics[width=0.8\linewidth]{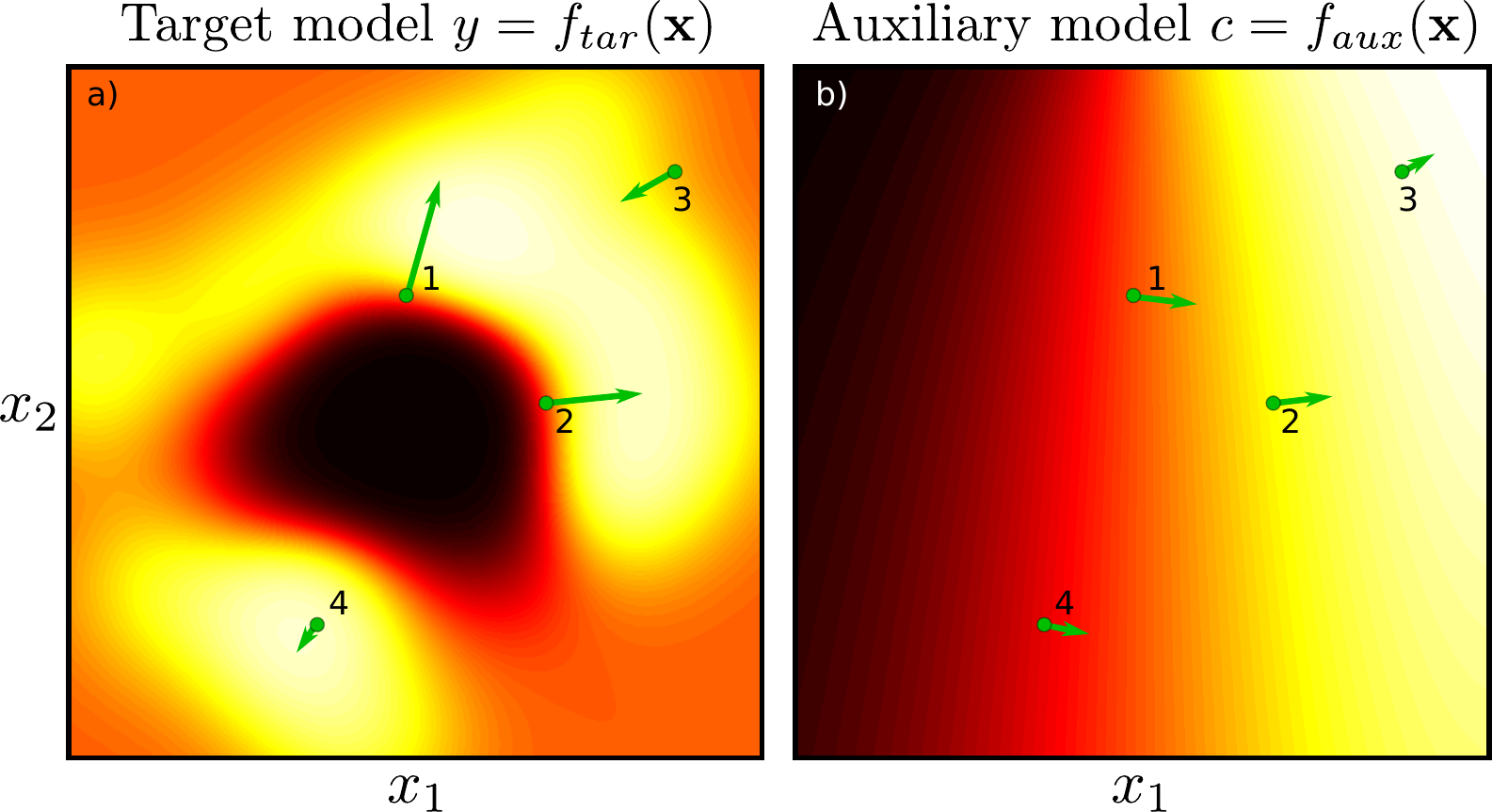}
    \caption{At the heart of our method is a simple idea;  if we adjust the model input so that the predicted protected variable changes, then the model output should not change. a) Target model predicts $y$ from inputs $\mathbf{x}$.  We wish to test fairness at points 1,2,3,4. b) We construct an auxiliary model that predicts protected variable $c$ from inputs $\mathbf{x}$.  fAux compares gradients  (green arrows) of the two models. Point 2 is unfair because the target and auxiliary model gradients are large and parallel; the model prediction changes as the protected variable changes. The other points are fair since the gradients are orthogonal (point 1) or one or other gradient is small (points~3,4). }
    \label{fig:faux_main}
    \vspace{-3mm}
\end{figure}

There exist methods to test models with all of these definitions.  \cite{agarwal2018automated} and \cite{galhotra2017fairness} use the FTU definition, \cite{john2020verifying} and \cite{wachter2018counterfactual} concentrate on the FTA definition, and \cite{black2019fliptest} use the CFF principle. However, each approach has limitations.  Features can act as surrogates of a protected variable, which FTU ignores. Using FTA needs domain-specific knowledge to define similarity metrics for inputs and outputs. Using CFF requires building a generative model to produce counterexamples. Moreover, we show experimentally that methods based on FTA can exhibit low precision.
%Moreover, we show experimentally that methods based on FTU exhibit low recall and those based on FTA exhibit low precision.
%Giuseppe: FTU is much more naive than this: the only variable it cares about is C specifically.  If C is not an input, a model satisfies FTU by definition.

In this paper, we introduce \textit{fAux} (pronounced fox), a new framework for individual fairness testing which avoids these difficulties.   Our contributions are as follows:  
%first, we show that all three of these fairness definitions can be unified in a projective and constraint formalism. 

\begin{enumerate}%[leftmargin=0pt, noitemsep]
    \item We propose a simple and general criterion for individual fairness testing: for a fair model, the derivative of the model prediction with respect to the protected variable should be small.
    \item We introduce the concept of an {\em auxiliary model} which describes the relationship between the input features and the protected variable and show how to use this to help evaluate this criterion.  Figure \ref{fig:faux_main} provides the intuition behind our approach.  We specifically focus on unfair treatment that is created by historical bias in datasets \cite{mehrabi2019survey}.
    \item To help evaluate fairness testing, we present a novel synthetic data generation method that merges multiple real datasets through a probabilistic graphical model to flexibly simulate realistic data with controllable bias levels.
\end{enumerate}

% we probably need to remove this problematic statement below, it actually do not matter since we simply doing test and it is effective. that is all we want to claim. -Wuga
%-----------
% Our method doesn't require explicit identification of features sensitive to the protected variable, nor does it require a generative model for the data. 
% In particular, the proposed method does not rely on similarity metrics specified by a domain expert. Instead, the metrics are derived in a principled fashion from a Local Independence Criterion.  We show that the performance of our approach is superior on both synthetic and real datasets, and is more computationally efficient than alternatives.
%----------
%\todo{Furthermore, the similarity metrics employed are not specified by an expert, but are derived in a principled fashion from a Local Independence Criterion.} % Try not write incomplete sentences..
%\Simon{This paragraph is essentially a sales pitch for the efficacy of the method.   Is this accurate?  Is there anything else that we can add?  Your original in comments below.  Note how I have changed the tone to make it sound more exciting!}

%In the experiment section, we compare the proposed method's performance and that of multiple state-of-the-art verification approaches on both 
%synthetic datasets and real datasets. We also record the inference time and consumption of computational resources to further justify the proposed %model's efficiency. Finally, we provided multiple local experiments to reveal the insight of the proposed model from different aspects.

\section{Preliminary Material}

%MODIFIED

%In this section, we introduce notation, review definitions of individual fairness and summarize their limitations. We, then, describe definition of fairness in a view that connects the earlier ones and forms the basis for a novel fairness test that we develop in this paper.

In this section, we describe a definition of individual fairness that unites a number of definitions in the literature.  We then analyze the technical challenges of existing tests, which serve as motivation for our own.  Throughout, we adopt the following notation:
\begin{itemize}
    \item $X$: Feature (input) variables. When features are observed, we use $\mathbf{x}$ to represent the feature vector.
    \item $Y$: Prediction (output) variables. When a label is observed, we use $y$ to represent the label as a scalar. As we do fairness testing on binary classification tasks, the prediction $y$ in this paper is a probability.
    \item $C$: Protected variables (e.g., gender). We use $\mathbf{c}$ to represent the observed values.
    \item $\phi$: Distance metric. $\phi_{in}(\cdot,\cdot)$ denotes a metric of input space, and $\phi_{out}(\cdot,\cdot)$ a metric of output space.
    \item $f_{tar}$: Target function for fairness testing. This takes features $\mathbf{x}$ as input and produces a prediction $\hat{y}$.
    \item $f_{aux}$: Auxiliary model. This takes features $\mathbf{x}$ as input and produces predictions $\mathbf{c}$ for protected attributes $C$.
\end{itemize}

\subsection{Individual Fairness Definitions}\label{sec:ind-fair-definition}

Individual fairness describes the tolerable discrepancy of model predictions at the level of individual data points.  There are several characterizations that operate at this level, and to facilitate their comparison we use the following definition.  Let the observed features $\mathbf{x}$ be generated from underlying latent variables $\mathbf{z}_{\bot}$ and $\mathbf{z}_{\parallel}$ via a function $f_g$:
\begin{equation}
    \mathbf{x} = f_g(\mathbf{z}_{\bot}, \mathbf{z}_{\parallel}).
\label{eq:factorize_features}
\end{equation}
Here, $\mathbf{z}_{\bot}$ denotes latent vectors with no correlation with the protected variables  $\mathbf{c}$, and $\mathbf{z}_{\parallel}$ is influenced by the protected variables $C$ through a function $\mathbf{z}_{\parallel} = \psi(\mathbf{c})$. 

\begin{definition}\label{def:canonical}
\label{df:canonical_individual_fairness}
A model $f_{tar}$ is individually fair, if it produces exactly identical outcomes when given input feature vectors $\mathbf{x}_i$ and $\mathbf{x}_j$ which share the same latent vector $\mathbf{z}_{\bot}$:
\begin{equation}
    f_{tar}(\mathbf{x}_i) = f_{tar}(\mathbf{x}_j),
\end{equation}
where $\mathbf{x}_{i} = f_{g}(\mathbf{z}_{\bot}, \psi(\mathbf{c}))$ and $\mathbf{x}_{j} = f_{g}(\mathbf{z}_{\bot}, \psi(\mathbf{c}^{\prime}))$.
\end{definition}

%%MODIFIED FROM ORIGINAL
%%ORIGINAL COPIED TO HERE:
%\input{related_work}

%\noindent Three families of fairness tests align with this procedure:  

At a high level, the definition states that given a pair of similar data points $\mathbf{x}_{i}$ and $\mathbf{x}_{j}$, they should receive equal treatment from a target model.  This perspective is shared by the three most common families of individual fairness tests:  \textbf{Fairness Through Unawareness}~(FTU) stipulates that protected variables should not explicitly be used in the prediction process.  ``Similar'' points are thus points that differ only in the protected variable, since the latter cannot affect the outputs. \textbf{Fairness Through Awareness}~(FTA) formally define ``similar'' points using a metric on the input space \cite{dwork2012fairness}. Finally, \textbf{Counterfactual Fairness}~(CFF) stipulates that the model predictions should not causally depend on protected variables $C$ \cite{CounterfactualFairness}.  For a given input $\mathbf{x}_i$, the ``similar'' individual $\mathbf{x}_j$ is the counterfactual constructed by intervening on $C$.  %Disparate treatment between these individuals measures the direct causal effect \cite{Zhang2016SituationTD}.

For these three tests the main technical challenge is the generation of ``similar'' pairs, either by searching within a neighbourhood defined by a metric, or by transformation of the input.  We now describe these challenges in more depth.

\subsubsection{Technical Challenges in Employing Similarity Metrics}

%Got rid of Lahoti_2019
%It's not experts labelling similarity, so much as who should receive similar treatment
The main challenge to constructing similarity metrics is the task-specific domain knowledge required~\cite{dwork2012fairness}.  Some methods simply employ unweighted $l_p$ norms \cite{wachter2018counterfactual,john2020verifying}. Others obtain weights from a linear model trained to predict the protected variable $c$ from the input $\mathbf{x}$ \cite{ruoss2020learning,yurochkin2020training}.  Others still learn metrics from expertly-labelled pairs \cite{mukherjee2020simple}.

%Modified:
%\cite{mukherjee2020simple,ilvento2020metric}.

\subsubsection{Technical Challenges in Transforming Inputs}

With regards to input transformations, the main challenges are stability, and the risk of generating out-of-distribution samples.  Both risks are faced by approaches that employ adversarial techniques~\cite{wachter2018counterfactual,maity2021statistical}.  When the dataset is small, and lower dimensional, Optimal Transport (OT) can be used to define a mapping between pairs of inputs based on pairwise-similarity~\cite{dwork2012fairness,pmlr-v97-gordaliza19a}.  However, this method scales poorly.  To this end, approximate OT methods, based on dual-formulations \cite{chiappa2021fairness} and Generative Adversarial Networks \cite{black2019fliptest} have also been explored - though again, these present risks of instability and out-of-distribution samples.

%Modified:
%\cite{wachter2018counterfactual,Ustun_2019,maity2021statistical}

CFF also employs generative models, specifically, causal graphical models.  However, such graphs are rarely available.  Moreover, training such models via unsupervised learning is hard \cite{ReducingModeCollapse}, increasing the risk that the generated inputs are out-of-domain, or have limited coverage.

%Remove: \cite{ImprovedGANTechniques,ReducingModeCollapse}

%Optimal transport only works with binary labels

\subsection{Mechanisms of Discrimination}\label{sec:mechanism-of-discrimination}

%MODIFIED FROM THE ORIGINAL
%REMOVED THESE TWO CITATIONS
%There are many reasons why models exhibit unfair behaviour, but one of the most common is the mishandling of {\em historical bias}~\cite{fuchs2018dangers,mehrabi2019survey,FrameworkForBias} 

We now describe the mechanism of discrimination our test aims to detect.  There are many reasons why models exhibit unfair behaviour, but one of the most insidious is the mishandling of {\em historical bias}~\cite{mehrabi2019survey}.  Here, pre-existing prejudices create spurious correlations between features $X$ and protected variables $C$\footnote{More in-depth discussion and worked examples may be found in Section \textit{Use Cases} of the supplementary materials.} (Figure~\ref{fig:pgm_data}). These correlations provide a model with two possible inference paths: one legitimate, and one that implicitly infers $C$.  The latter paths render the FTU definition unreliable, as models learn to exploit surrogate features even when $C$ is omitted~\cite{FairMLBook}.  By contrast, depending on the metric, FTA is sensitive to all variations in the input, and thus may flag instances where inference was legitimate.  

In the next section, we present an approach that can precisely distinguish between these two paths.  It does so scalably and operates within distribution. Moreover, there is lower overhead, as it requires no domain-expertise, and it employs only supervised learning techniques.

%%MODIFIED FROM THE ORIGINAL: REMOVED
% While the latent variables or generative functions may not be observable/recognizable in general, the above definition holds in practice whenever $\mathbf{x}$ can be partially influenced by protected variables $\mathbf{c}$.

%%MODIFIED FROM THE ORIGINAL: REMOVED
%\input{connect_earlier_notions_of_ind_fairness}

\section{Local Fairness Tests with fAux}\label{sec:local_fairness_test}

\begin{figure}[t]
     \centering
     \begin{subfigure}[b]{0.32\linewidth}
     \centering
        \includegraphics[width=0.8\textwidth]{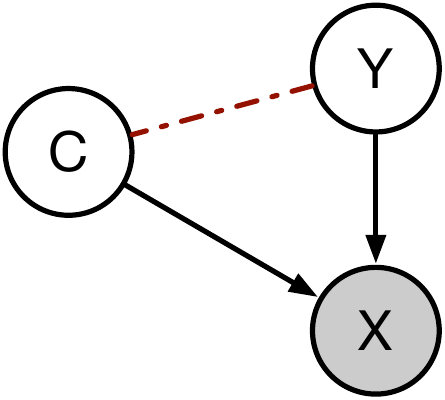}
        \caption{Data Generation}
        \label{fig:pgm_data}
     \end{subfigure}
     \hspace{\fill}
     \begin{subfigure}[b]{0.32\linewidth}
     \centering
        \includegraphics[width=0.8\textwidth]{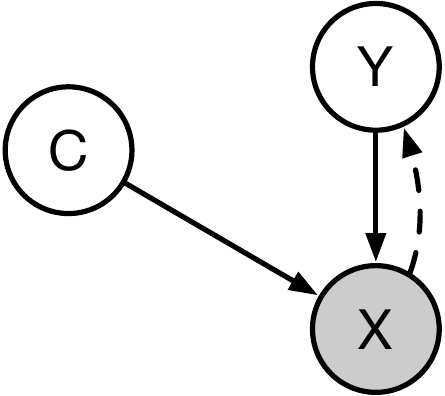}
        \caption{Fair Model}
        \label{fig:pgm_fair}
     \end{subfigure}
     \hspace{\fill}
    \begin{subfigure}[b]{0.32\linewidth}
    \centering
        \includegraphics[width=0.8\textwidth]{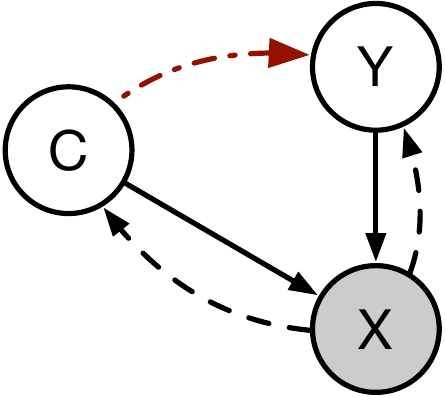}
        \caption{Unfair Model}
        \label{fig:pgm_unfair}
     \end{subfigure}
        \caption{\textbf{Graphical models to show how historical bias causes unfairness.} (a) Generation process of a biased training dataset. Red dashed line denotes that $Y$ and $C$ may have correlations due to historical bias. (b) Fair models learn to infer $Y$ while cancelling the impact from protected variables $C$. Solid arrows show generative dependence. Dashed arrow shows learned inference mapping. (c) Unfair models infer  protected variables to support their predictions.}
        \label{fig:graphical-models}
\vspace{-5mm}
\end{figure}

We now develop a novel fairness testing method that satisfies Definition \ref{def:canonical}, but does not suffer from the limitations described previously.  We start by using the graphical model in Figure \ref{fig:graphical-models} to propose a criterion for individual fairness based on sensitivity analysis.    We use this to motivate the {\em Local Independence Criterion (LIC)} which examines whether a model suffers from historical bias, and show that this satisfies Definition \ref{def:canonical}. Finally, we introduce the idea of an {\em Auxiliary Model} which is needed to create a practical test.

% \subsection{Motivation}
% \label{sec:motivation}
% \vspace{-2mm}

% Figure \ref{fig:graphical-models} suggests that for a model to be fair, then conditional independence should hold so $
% P(y,\mathbf{c}|\mathbf{x}) = P(y|\mathbf{x})P(\mathbf{c}|\mathbf{x})$,
% %\Simon{I changed this to a single line to save space and also because this is not the normal way that conditional independence is written, but put if back if you feels strongly}
% for all $y\in Y$, $\mathbf{x} \in X$, and $\mathbf{c}\in C$. In other words, having observed data $\mathbf{x}$ the posterior distribution $P(y|\mathbf{x})$ contains no information about the posterior distribution $P(\mathbf{c}|\mathbf{x})$ over the protected variable. In the presence of noise, we do not expect this relation to exactly hold and so we relax this constraint using a logarithmic form with associated  tolerance $\delta$ such that:
% \begin{equation}
%     \sup_{\mathbf{c}}\left|\log P(y|\mathbf{c},\mathbf{x}) - \log P(y|\mathbf{x})\right| \leq \delta
% \label{eq:prob_diff}
% \end{equation}
% where $\mathbf{c}$ is any feasible value vector in the domain of protected variables $C$.

\subsection{Local Independence Criterion}
\label{sec:differentiation}
\label{sec:lic}
%Now, we present a novel approach to flag discriminatory model behaviour at the individual level by introducing a Local Fairness Test~(LFT).
A sufficient condition for a model to violate the individual fairness criterion of Definition \ref{def:canonical} is that its prediction depends on the protected variable $\mathbf{c}$. We may reveal this dependence locally using the partial derivative
\begin{equation}
    \frac{\partial f_{tar}(\mathbf{x})}{\partial \mathbf{c}} \neq 0,
\end{equation}
which indicates that the prediction is sensitive to a small perturbation of $\mathbf{c}$. In practice, the machine learning model and data introduce inevitable noise, thus we relax the above expression with a pre-defined tolerance $\delta$.  We then obtain:

\begin{theorem}\label{theorem:LIC} Let there exist a generative model $f_g$ that influences features $X$ with protected variables $C$, such that $\mathbf{x} = f_g(\mathbf{z}_{\bot}, \psi(\mathbf{c}))$.
If a machine learning model $f_{tar}$ violates the \textit{Local Independence Criterion (LIC)} 
\begin{equation}
\begin{aligned}
%\max_{k}
\left|\frac{\partial f_{tar}( \mathbf{x})}{\partial \mathbf{c}}\right|_{\infty}
    %&= \lim_{\Delta{\mathbf{c}}\to 0} \frac{f_{tar}(f_g(\mathbf{z}_{\bot}, \psi(\mathbf{c}))) - f_{tar}(f_g(\mathbf{z}_{\bot}, \psi(\mathbf{c}+\Delta\mathbf{c})))}{\Delta{\mathbf{c}}}\\
    %\leq \lim_{\Delta{\mathbf{c}}\to 0}\frac{\delta}{\Delta{\mathbf{c}}},
    \leq \delta,
\end{aligned}
\label{eq:local_independent_definition}
\end{equation}
by a pre-defined threshold $\delta$, then the model $f_{tar}$ violates the individual fairness criteria in Definition~\ref{df:canonical_individual_fairness}.
%~\ref{sec:theorem_proof}.
\end{theorem}
%\Simon{Wuga!  I need some help here.  I'm not sure that the equation number here is correct.  Can you check?}
%sure on it. fixed
% \Wuga{such that
%\begin{equation}
%\begin{aligned}
%    \int_\mathbf{c} \left|\nabla f_{tar}\frac{\partial \mathbf{x}}{\partial \mathbf{c}}\right|dc &\leq \sum_{\Delta\mathbf{c}}\delta P(\Delta\mathbf{c})\\
%    |f_{tar}(\mathbf{x}_i) - f_{tar}(\mathbf{x}_j)| &\leq \delta
%\end{aligned}
%\end{equation}
%where the right hand side falls back to summation as the integration is on a probability mass.}
%\Christopher{This proof should be added to the appendix}
% Wuga: agreed

\noindent A proof is provided in Appendix \ref{sec:theorem_proof}, but intuitively, the partial derivative considers the disparate treatment of two infinitesimally-close individuals.

To use the LIC, we need to estimate the derivative $\partial f_{tar}( \mathbf{x})/\partial \mathbf{c}$, for which the chain rule gives:
\begin{equation}
    %\max_k
    \left|\frac{\partial f_{tar}( \mathbf{x})}{\partial \mathbf{x}} \frac{\partial \mathbf{x}}{\partial \mathbf{c}}\right|_{\infty} \leq
    \delta.
    %\lim_{\Delta{\mathbf{c}}\to 0}\frac{\delta}{\Delta{\mathbf{c}}},
\label{eq:chain_rule_expansion_lic}
\end{equation}
Unfortunately, the term $\partial \mathbf{x}/\partial \mathbf{c}$ is undefined without accessing the underlying generative model that maps protected variables $C$ to the features $X$ and this is rarely available\footnote{Note, the protected variables $C$ are not necessarily continuous as we will model the mapping through an auxiliary model later. }.

\subsection{Auxiliary Models}\label{sec:auxiliary}

In this section we suggest an approximation of $\partial \mathbf{x}/\partial \mathbf{c}$ that requires neither generative model nor attempts to model the latent representations $\mathbf{z} \in Z$. 

One approach would be to build a model to predict $\mathbf{x}$ from $\mathbf{c}$ and use the derivative of this model to approximate $\partial \mathbf{x}/\partial \mathbf{c}$.  However, this might be poor as the number of protected variables is often far smaller than the feature size. Instead, we consider the inverse problem:  we describe the mapping from features $X$ to the protected variables $C$ using an \textit{auxiliary model} $\mathbf{c}=f_{aux}(\mathbf{x})$. We then invert this model in a local neighbourhood around a given point $\mathbf{x}_0$, to approximate the desired derivative. To this end, given this auxiliary model $f_{aux}$  we apply the Taylor expansion around  $(\mathbf{x}_0,\mathbf{c}_0)$:
\begin{equation}
    \mathbf{c} - f_{aux}(\mathbf{x}_0) \approx \left(\frac{\partial f_{aux}(\mathbf{x}_0)}{\partial \mathbf{x}_0}\right)^{\top}(\mathbf{x} - \mathbf{x}_0) \label{eq:m-taylor-expansion}
\end{equation}
where we replaced $\mathbf{c}_0$ with its prediction $f_{aux}(\mathbf{x}_0)$.
The left-hand side denotes the change in the space of protected variables. The right hand side is a Jacobian vector product. 
%\Christopher{This equation 10 needs more explanation. How is the right hand side a dot product? I only see the term related to $f_{aux} $. I like the idea of explaining the left and right hand sides before going further but it needs to be more clear what each one means. I feel this is a critical point in the paper. The explanations about inverting the model up to here were clear but this is the first point of confusion.}
% wuga: addressed somehow, but this one need to discuss during meeting.  

We apply the Moore-Penrose pseudo-inverse to find the minimum norm solution for $\mathbf{x}$: 
\begin{equation}
    f_{aux}^{-1}(\mathbf{c}) =\mathbf{x}_0 +  (\mathbf{c} - f_{aux}(\mathbf{x}_0))\left(\nabla f_{aux}^\top\nabla f_{aux}\right)^{-1}\nabla f_{aux}^\top %\nonumber 
    \label{eq:penrose-invers}
\end{equation}
where we use $\nabla f_{aux}$ to denote $\partial f_{aux}(\mathbf{x}_0)/\partial \mathbf{x}_0$. This allows us to approximate $\partial \mathbf{x}/\partial \mathbf{c}$ by
\begin{equation}
    \frac{\partial \mathbf{x}}{\partial \mathbf{c}} \approx \frac{\partial f_{aux}^{-1}(\mathbf{c})}{\partial \mathbf{c}} = \left(\nabla f_{aux}^\top\nabla f_{aux}\right)^{-1}\nabla f_{aux}^\top.
\label{eq:derivative_approximation}
\end{equation}
Finally, by combining Equation~\ref{eq:derivative_approximation} and the Equation~\ref{eq:chain_rule_expansion_lic}, we can approximate the LIC with:
\begin{equation}
\left|\nabla f_{tar} \left(\nabla f_{aux}^\top\nabla f_{aux}\right)^{-1}\nabla f_{aux}^\top \right|_{\infty} \leq \delta
\label{eq:grad-alignment}
\end{equation}
where we use $\nabla f_{tar}$ to denote $\partial f_{tar}(\mathbf{x}_0)/\partial \mathbf{x}_0$.

%Because we are evaluating a partial derivative, we need only examine perturbations that maximize the change in $c$, but leave $z$ and $y$ approximately constant. These perturbations may be found using a surrogate model. 

% \todo{The approaches used in [][] also employ an auxiliary model to define their metric, except this model is linear (and therefore the metric is uniform).  By contrast, our model is non-linear, and therefore non-uniform, and in our experiments we show there is a benefit to this higher expressivity.}

%fAux is also distinct in how it solves problem (2).  

%\todo{Unlike ..., our test can flag individual discrimination}

% \begin{enumerate}
%     \item The choice of fairness metric $\phi_{in}$
%     \begin{itemize}

%     \end{itemize}
%     \item An algorithm for selecting pairs of points $(\textbf{x}_i, \textbf{x}_j)$ given $\phi_{in}$.
%     \begin{itemize}
%         \item Based on optimal transport
%         \item Based on adversarial attacks
%     \end{itemize}
% \end{enumerate}

\subsection{On the Choice of the Auxiliary Model}

In this paper, we employ Multilayer Perceptron (MLP) architectures for our auxiliary models, to minimize the amount of model overhead.  Though inverting such models can potentially yield low-fidelity reconstructions of $\mathbf{x}$, additional fidelity may require modelling factors of $\mathbf{x}$ that are independent of $\mathbf{c}$.  The LIC avoids the need for these additional factors since the end goal is estimating the partial derivative of $\frac{\partial \mathbf{x}}{\partial c}$ only (for more details, see Appendix \ref{sec:pseudo}).  In our experiments, we demonstrate that even such simple architectures are sufficient to achieve state of the art results.  

Beyond their simplicity, another advantage of using MLPs is their flexibility, as they easily accommodate both real-valued and categorical outputs.  This makes it possible to analyze scenarios in which the protected variable $C$ is not binary.

%What's more, we note that, in comparison with generative modelling, 

\subsection{Relaxations and Extensions}\label{sec:relaxations-and-extensions}
While the basic fAux described above is sufficient for an individual fairness test, it depends heavily on the behaviour of $\nabla f_{tar}$, which may be ill-conditioned.  In this section, we introduce several variants of the basic fAux method.

% \noindent {\bf Sign of Gradient~(fAux+SG):} As the gradient of auxiliary model $f_{aux}$ may produce large variance, we propose using the sign of the gradient to stabilize the testing. Specifically, instead of using Equation~\ref{eq:grad-alignment}, we propose the alternative:
% \begin{equation}
% \left|\nabla f_{tar}\; \mathrm{sign}(\nabla f_{aux}^{\top})\right|_{\infty} \leq \delta
% \label{eq:sign-alignment}
% \end{equation}

\noindent {\bf Normalization of Gradient~(fAux+NG):} Different features may have very different valid ranges and so the gradient of either the target or the auxiliary model could be biased towards a subset of features. To mitigate this problem, we use an $l_2$ normalization of the gradients to give the criterion:
\begin{equation}
\left|\mathrm{norm}(\nabla f_{tar}) \; \mathrm{norm}(\nabla f_{aux}^{\top})\right|_{\infty} \leq \delta.
\label{eq:norm-alignment}
\end{equation}
where we removed the inverse term as this normalization is longer needed.

\noindent {\bf Integrated Gradient~(fAux+IG):} We substitute the raw gradients $\nabla f_{tar}$ and $\nabla f_{aux}$, for integrated gradients~\cite{sundararajan2017axiomatic} that provide a smoothed gradient signal.

\section{Experiments}\label{sec:experiments}
% \Christopher{General comment: Is it fAux or fAux?, section 3.2.1 also alternates between the two acronyms}
% wuga: addressed

We now evaluate the proposed fAux test to answer the following research questions~(RQs):
\begin{itemize}[leftmargin=0pt, noitemsep, label={}]
    \item {\bf RQ1:} Given the target model $f_{tar}$ trained on synthetic datasets whose ground-truth degrees of bias are known, how well does fAux perform compared to other testing methods?
    \item {\bf RQ2:} Given the target model $f_{tar}$ trained on the real dataset whose ground-truth degree of bias is unknown, can fAux identify discriminatory features?
    %\Christopher{What do you mean by expensive analytical method? Are you trying to say that these discriminatory features can be identified more efficiently than when using the other methods? If so can you rephrase to say this?}
    % wuga: just confirmed with giuseppe. those groudtruth are given in other papers
    \item {\bf RQ3:} How efficient is fAux compared to the existing approaches in terms of inference cost?
    \item {\bf RQ4:} Does fAux have any conditions needed to guarantee reliable test performance? In particular, we want to know how the effect of auxiliary model performance would impact the test performance.
\end{itemize}

% experiment section is very objective, no contribution highlight. only say the facts: what you did, what you observed, where is the evidence.
%The main contribution of this paper is an individual fairness test, which flags discrimination with high precision, but incurs minimal modelling overhead.  To evaluate our approach, we require ground truth labels for instances of individual discrimination by a target model.  In Section \ref{sec:synth-data-overview}, we describe how to implement synthetic datasets with the structure of Figure \ref{fig:graphical-models}, allowing us to generate ground truth labels in a tuneable setting.  We then employ these labels in Section \ref{sec:benchmark-tests} to benchmark the precision of our local fairness tests against existing approaches.  Finally, in Section \ref{sec:real-dset-tests}, we deploy our local fairness test to identify mechanisms of discrimination provoked by real datasets.

\subsection{Experimental Setup}
%In this sub-section, we describe our experiment settings, including the target model architecture, the baseline algorithm descriptions, data descriptions as well as the metrics used for the evaluations.

\subsubsection{Candidate Testing Algorithms and Target Model}

\begin{itemize}[leftmargin=0pt, label={}]
\item {\bf FTA:} A local version of FTA, inspired by works such as \cite{john2020verifying}. In these approaches, an $\epsilon$-neighbourhood around the input $x_i$ is rigorously searched over to bound the output deviation $\phi_{out}(f_{tar}(\textbf{x}_i), f_{tar}(\textbf{x}_j))$.  In the limit that $\epsilon$ goes to zero, this is equivalent to a bound on the $l_p$ norm of $\nabla f_{tar}$.  We also construct a weighted $l_p$-norm using a linear auxiliary model, as is common in the individual fairness literature \cite{ruoss2020learning,yurochkin2020training}.

%using an unweighted $l_2$-norm in the input space.  With this choice of metric the test statistic becomes the norm of the target model gradient $|\nabla f_{tar}|_2$.  

\item {\bf Unfair Map:} \cite{maity2021statistical} uses a gradient-flow attack to generate pairs of points that violate FTA.  This attack is conducted within a neighbourhood defined by a similarity metric $\phi_{in}$, and to this end we employ the same weighted $l_p$ norm used in FTA.
%The test statistic is then the ratio between the loss function on these pairs. 

\item {\bf FlipTest:}~\cite{black2019fliptest} approximates CFF by leveraging Wasserstein GANs to generate pairs of inputs.  %We use fully connected networks for both the generator and the discriminator.

% We use MLPs of varying depth and width for our Generators and Discriminators.  Due to the sensitive nature of GAN training, we tune the batch size, in addition to the following hyperparameters:
% \begin{enumerate}
%     \item $n_{critic}$, which controls the relative number of training steps between the Dicriminator and the Generator
%     \item $\lambda_{c}$, the weight of the transport cost in the Generator loss function specific to \cite{FlipTest}.
% \end{enumerate}

%\item {\bf fAux:} We include all variants of the fAux method proposed in this paper and described in the Section \ref{sec:relaxations-and-extensions}. %We set the auxiliary model architecture to be a fully connected network.  

% don't say anything about data generation process in the algorithm description.
%Knowing the form of the fusion function $F$, we constrain the input layers of the auxiliary models to project out any non-$C$ dependence in the dataset.

\item \textbf{LIC-UB:} An upper-bound on the Local Independence Criterion~(LIC).  For the experiments on synthetic datasets, we can compute the true gradient of the generative model to conduct the LIC check \eqref{eq:chain_rule_expansion_lic}. Since error introduced by the approximation is removed, the test performance should achieve its upper-bound.

%\item \textbf{LIC-UB:} The upper-bound performance of a testing approach that aims to enforce the Local Independence Criterion~(LIC) described previously. For the experiments on synthetic datasets, we can access the ground truth generative model for the data and so we can compute the true gradient of the generative model (instead of approximating it with the auxiliary model) to conduct the LIC check. Since error introduced by the approximation is removed, the test performance should achieve its upper-bound.

\item \textbf{Target Models:} Given datasets in the form $D = \{\cdots(\mathbf{x}, y, \mathbf{c})\cdots\}$, we train multi-layer fully connected networks as the target models $f_{tar}$ with only features $\mathbf{x}$ and label $y$. The target models in our experiments are all classifiers that aim to produce probabilistic predictions $P(y|\mathbf{x})$. However, as previously discussed, an unfair model may infer protected variable $C$, resulting in it implicitly modeling $P(y|\mathbf{x},\mathbf{c})$.
\end{itemize}

\begin{table*}[t]
\centering

\resizebox{\linewidth}{!}{%
\begin{tabular}{l|c|c|c|c|c|c|c|c}
\toprule
& \multicolumn{1}{c|}{Synthetic-1}                                & \multicolumn{1}{c|}{Synthetic-2}                                & \multicolumn{1}{c|}{Synthetic-3}                                & \multicolumn{1}{c|}{Synthetic-4}                               & \multicolumn{1}{c|}{Synthetic-5}                                   & \multicolumn{1}{c|}{Synthetic-6}                                   & \multicolumn{1}{c|}{Synthetic-7}                                   & \multicolumn{1}{c}{Synthetic-8}                                   \\ \midrule \midrule
xc/xy                                                             & \begin{tabular}[c]{@{}c@{}}magic/\\ backache\end{tabular} & \begin{tabular}[c]{@{}c@{}}magic/\\ backache\end{tabular} & \begin{tabular}[c]{@{}c@{}}magic/\\ backache\end{tabular} & \begin{tabular}[c]{@{}c@{}}magic/\\ backache\end{tabular} & \begin{tabular}[c]{@{}c@{}}australian/\\ credit\end{tabular} & \begin{tabular}[c]{@{}c@{}}australian/\\ credit\end{tabular} & \begin{tabular}[c]{@{}c@{}}australian/\\ credit\end{tabular} & \begin{tabular}[c]{@{}c@{}}australian/\\ credit\end{tabular}\\ \midrule
Bias Level  & 0.5  & 0.5  & 1  & 1  & 0.5  & 0.5  & 1  & 1  \\ \midrule 
Fusion Approach  
& outer  & concat  & outer  & concat  & outer  & concat  & outer & concat        
\\ \midrule \midrule
FTA  &
\begin{tabular}[c]{@{}c@{}} 0.307 $\pm$ 0.000 \end{tabular} &
\begin{tabular}[c]{@{}c@{}} 0.306 $\pm$ 0.000 \end{tabular} &
\begin{tabular}[c]{@{}c@{}} 0.307 $\pm$ 0.000 \end{tabular} &
\begin{tabular}[c]{@{}c@{}} 0.307 $\pm$ 0.000 \end{tabular} &
\begin{tabular}[c]{@{}c@{}} 0.466 $\pm$ 0.000 \end{tabular} &
\begin{tabular}[c]{@{}c@{}} 0.293 $\pm$ 0.000 \end{tabular} &
\begin{tabular}[c]{@{}c@{}} 0.355 $\pm$ 0.000 \end{tabular} &
\begin{tabular}[c]{@{}c@{}} 0.309 $\pm$ 0.000 \end{tabular} 
\\ \midrule
FTA + lin. aux
& 0.402 $\pm$ 0.000
& 0.683 $\pm$ 0.003	
& 0.451 $\pm$ 0.001	
& 0.455 $\pm$ 0.001
& 0.709 $\pm$ 0.000	
& 0.658 $\pm$ 0.00	
& 0.672 $\pm$ 0.000	
& 0.612 $\pm$ 0.000
\\ \midrule
Unfair Map
& 0.407 $\pm$ 0.000	
& 0.400 $\pm$ 0.000
& 0.576 $\pm$ 0.000	
& 0.820 $\pm$ 0.000
& 0.571 $\pm$ 0.000	
& 0.653 $\pm$ 0.000
& 0.687 $\pm$ 0.000
& 0.732 $\pm$ 0.000
\\ \midrule
FlipTest
& 0.598 $\pm$ 0.285 
& 0.880 $\pm$ 0.015
& 0.433 $\pm$ 0.259
& 0.633 $\pm$ 0.109 
& 0.600 $\pm$ 0.148
& 0.678 $\pm$ 0.056  
& 0.701 $\pm$ 0.209 
& 0.732 $\pm$ 0.132 
\\ \midrule  \midrule
fAux                                                        & \begin{tabular}[c]{@{}c@{}} 0.332 $\pm$ 0.004 \end{tabular} &
\begin{tabular}[c]{@{}c@{}} 0.998 $\pm$ 0.001 \end{tabular} &
\begin{tabular}[c]{@{}c@{}} 0.311 $\pm$ 0.002 \end{tabular} &
\begin{tabular}[c]{@{}c@{}} \textbf{1.000 $\pm$ 0.000} \end{tabular} &
\begin{tabular}[c]{@{}c@{}} 0.615 $\pm$ 0.032 \end{tabular} &
\begin{tabular}[c]{@{}c@{}} 0.937 $\pm$ 0.005 \end{tabular} &
\begin{tabular}[c]{@{}c@{}} 0.564 $\pm$ 0.014 \end{tabular} &
\begin{tabular}[c]{@{}c@{}} 0.997 $\pm$ 0.001 \end{tabular}   
%\\ \midrule
%\begin{tabular}[c]{@{}l@{}}fAux+SG\end{tabular}      & \begin{tabular}[c]{@{}c@{}}0.750 $\pm$ 0.048\end{tabular}  & \begin{tabular}[c]{@{}c@{}}0.715 $\pm$ 0.034\end{tabular} & \begin{tabular}[c]{@{}c@{}}0.822 $\pm$ 0.021\end{tabular} & \begin{tabular}[c]{@{}c@{}}0.742 $\pm$ 0.076\end{tabular} & \begin{tabular}[c]{@{}c@{}}0.805 $\pm$ 0.020\end{tabular}    & \begin{tabular}[c]{@{}c@{}}0.808 $\pm$ 0.022\end{tabular}    & \begin{tabular}[c]{@{}c@{}}0.955 $\pm$ 0.039\end{tabular}    & \begin{tabular}[c]{@{}c@{}}0.910 $\pm$ 0.037\end{tabular}     
\\ \midrule 
\begin{tabular}[c]{@{}l@{}}fAux+NG\end{tabular}      & 
\begin{tabular}[c]{@{}c@{}} \textbf{0.876 $\pm$ 0.015} \end{tabular} &
\begin{tabular}[c]{@{}c@{}} \textbf{0.999 $\pm$ 0.001} \end{tabular} &
\begin{tabular}[c]{@{}c@{}} \textbf{0.978 $\pm$ 0.023} \end{tabular} &
\begin{tabular}[c]{@{}c@{}} \textbf{1.000 $\pm$ 0.000} \end{tabular} &
\begin{tabular}[c]{@{}c@{}} 0.815 $\pm$ 0.015 \end{tabular} &
\begin{tabular}[c]{@{}c@{}} \textbf{0.947 $\pm$ 0.006} \end{tabular} &
\begin{tabular}[c]{@{}c@{}} 0.910 $\pm$ 0.019 \end{tabular} &
\begin{tabular}[c]{@{}c@{}} 0.998 $\pm$ 0.001 \end{tabular} 
\\ \midrule
\begin{tabular}[c]{@{}l@{}}fAux+IG\end{tabular} & 
\begin{tabular}[c]{@{}c@{}} 0.717 $\pm$ 0.031 \end{tabular} &
\begin{tabular}[c]{@{}c@{}} 0.998 $\pm$ 0.000 \end{tabular} &
\begin{tabular}[c]{@{}c@{}} 0.937 $\pm$ 0.012 \end{tabular} &
\begin{tabular}[c]{@{}c@{}} \textbf{1.000 $\pm$ 0.000} \end{tabular} &
\begin{tabular}[c]{@{}c@{}} \textbf{0.886 $\pm$ 0.015} \end{tabular} &
\begin{tabular}[c]{@{}c@{}} 0.944 $\pm$ 0.002 \end{tabular} &
\begin{tabular}[c]{@{}c@{}} \textbf{0.979 $\pm$ 0.007} \end{tabular} &
\begin{tabular}[c]{@{}c@{}} \textbf{0.999 $\pm$ 0.001} \end{tabular} 
 
\\ \midrule
LIC-UB & 
\begin{tabular}[c]{@{}c@{}} 0.998 \end{tabular} &
\begin{tabular}[c]{@{}c@{}} 0.999 \end{tabular} &
\begin{tabular}[c]{@{}c@{}} 1.000 \end{tabular} &
\begin{tabular}[c]{@{}c@{}} 1.000 \end{tabular} &
\begin{tabular}[c]{@{}c@{}} 0.965 \end{tabular} &
\begin{tabular}[c]{@{}c@{}} 0.966 \end{tabular} &
\begin{tabular}[c]{@{}c@{}} 0.999 \end{tabular} &
\begin{tabular}[c]{@{}c@{}} 0.999 \end{tabular} 
\\
\bottomrule
\end{tabular}
}
\caption{\textbf{Performance Comparison among Individual Fairness Testing Methods on Synthetic Datasets} We report Average-Precision scores with the highest score in bold font. The confidence interval comes from 10 runs by re-training auxiliary models.  Rows are sorted in order of increasing computational requirements.  The last row corresponds to a theoretical upper bound.}
\label{table:synth-data-res}
\vspace{-3mm}
\end{table*}

\noindent Full implementation details are found in Appendix \ref{sec:hyperparams-for-tests}. %We here describe the different fairness tests 

\subsubsection{Synthetic Datasets with Ground Truth Bias}\label{sec:exp:synthetic-exp}
% outlines
% 1. Why synthetic by fusing (manually defined synthetic data is not complex enough to simulate the reality) STOCASITY AND NONLINEARITIES
% Generating synthetic datasets via graphical models allows us to compare the performance of the test methods with ground truth labels. However, we note that the existing synthetic generation approaches~(see \cite{kim2019learning}) are inadequate to simulate complex historical bias in the real-world datasets due to their limited stochasticity and nonlinearity. 
% \Simon{This worries me.  We are criticizing someone else's work without really giving much explanation or justification.  Has someone else made this point? If so we should cite.  If not, then we should elaborate.}
% Yes, we need to elaborate this. Too aggressive to say this. Smoothed wording. maybe less aggressive now? -Wuga
In producing our synthetic datasets, our goal was to retain the noisy and nonlinear relationships that are present in real datasets.  To this end, we construct a pipeline which joins real, unbiased datasets together via \textit{fusion} operations, based on a intentionally biased data sampling process.  Specifically, given two datasets $\hat{D}=\{\cdots (\hat{\mathbf{x}}_i,\hat{y}_i)\cdots\}$ and $\tilde{D}=\{\cdots (\tilde{\mathbf{x}}_j,\tilde{y}_j)\cdots\}$, and a fusion operation $f_{fus}$, we can produce a synthetic dataset $D_{syn}$ such that
   $D_{syn} = \{\cdots(\mathbf{x}, y, c)\cdots\} = \{\cdots (f_{fus}(\hat{\mathbf{x}}_i,\tilde{\mathbf{x}}_j), \hat{y}_i, \tilde{y}_j)\cdots\}$,
where $y\stackrel{\text{def}}{=}\hat{y}_i$, $c\stackrel{\text{def}}{=}\tilde{y}_j$, and $f_{fus}$ is a fusion operation (see below for examples). While this looks simple, the selection of data indices $i$ and $j$ for fusion is based on the predefined generative model under the hood. Furthermore, the generative model controls the degree of bias for the synthetic datasets with hyper-parameters. Thus, the data generation process reproduces the historical bias  described in Section \ref{sec:mechanism-of-discrimination}.  
%%%\Simon{This section still needs some work, I think. I don't understand equation 20. How does $c$ relate to the fusion operation?  It is on the LHS but not the RHS.  }
% yes, this should be fixed. as it merges two datasets, each of them would have a label variable, so one of the variable would be the target variable we want the target model to learn and the another label is pretended to be the protected variable c.  so c -> y_j, added extended explanation under equation

% it merges two sets of features into one vector through some functions. maybe fusion function is more straightforward.
% What does merge mean?  Combine on a per-element level?
% there are two options giuseppe suggested. one is is simple concatenation, another is abit complex through outproduct + flattening

% Okay.. just read about those below.

% 3. Hyper-parameters to control the complexity of the biased synthetic data, which we will use in tables.
A full description of the synethetic data pipeline~(including generative model specifics) can be found in Appendix \ref{sec:synthetic_data_generation}. Here, we summarize two key hyper-parameters of the data generator, which we will use to control ground-truth data bias and complexity.
\begin{itemize}[leftmargin=0pt, noitemsep, label={}]
    \item {\bf Bias Level:} The bias level controls the level of dependency between $Y$ and $C$ in the range of $[0,1]$. A Higher bias level results in larger correlation between $Y$ and $C$ in the generated dataset.
    \item {\bf Fusion Function:} The fusion function merges feature vectors from the two datasets.  We have two variants: 
    {\em Concatenation} which stacks features without changing the element values~(see \cite{CounterfactualFairness}), %This mimics the scenario described in \cite{CounterfactualFairness}, in which observable features may be partitioned into descendents and non-descendents of $C$. 
    and the {\em outer product} which blends features perfectly.
\end{itemize}

% 4. With deep generative model, we have ground truth bias for each validation data points by directly computing Canonical Fairness Score for the given target model for each data point. 

\subsubsection{Evaluating Fairness Tests and Selecting $\delta$}

Having access to the ground-truth generative model, we can compute the individual fairness score~(IFS) described in Definition~\ref{df:canonical_individual_fairness} for each generated synthetic data sample. IFS will serve as the ground-truth label in the following experiments on synthetic datasets.

% In flagging discrimination in practice, it is necessary to set a value for the threshold $\delta$.  This threshold is usually set by regulatory standards that depend on problem domain or statistics of manual auditing results.  Indeed, the selection of specific thresholds is a subtle question \cite{corbettdavies2018measure}, and in practice, decision-making may involve multiple thresholds (one or more thresholds for each population) that require further domain-specific study \cite{Corbett-DaviesP17}.

In flagging discrimination in practice, it is necessary to set a value for the threshold $\delta$.  This threshold is usually set by regulatory standards that depend on problem domain or statistics of manual auditing results.  Indeed, the selection of specific thresholds is a subtle question, and in practice, decision-making may involve multiple thresholds that require further domain-specific study \cite{corbettdavies2018measure}. %Omitted \cite{Corbett-DaviesP17}.

Thus, rather than choose a particular threshold $\delta$, since each candidate test yields a continuous discrimination score, we may construct a precision-recall curve.  We then compare the tests based on average precision. This enables us to determine which test is the most reliable across a range of different thresholds.

% Each of these tests outputs a continuous discrimination score, and may be thresholded by a cutoff $\delta$, to obtain a binary discrimination label.  Accordingly, we may combine this score with the ground truth labels provided by our synthetic datasets to compute the average precision score of the different tests.  The score for each test is computed across a series of target models that have been trained with different parameters and constraints.  The tests for FlipTest and fAux involve different hyperparameters, and we show the mean, variance, and maximum scores obtained in Table \ref{table:synth-data-res}.  We restrict our attention to the disadvantaged population, defined by $c = 0$.
%\Simon{We no longer use the word canonical.  We might need something else here if you have a good name for it}
% fixed now IFS 

\subsubsection{Real Datasets}\label{sec:real-dset}
We also compare the proposed model with the baseline models on two real datasets (and one more in the Appendix): 
\begin{itemize}[leftmargin=0pt, label={}]
    % \item {\bf Chicago SSL Dataset \cite{chicago_2017}:} The protected variable a binarized race variable (Black/White). %Here, we adopt the target model that are used in \cite{black2019fliptest}' work.
    \item {\bf Adult Income Dataset \cite{Dua:2019}:} The protected variable is gender, a binary variable (Female/Male). %Here the target model is a simple logistic regression model.
    % Our target model is a simple logistic regressor, which we design using a correlation analysis similar to \cite{FlipTest}.  We use an MLP for the auxiliary model, and one-hot encode the categorical features.  The total score for a categorical feature is the sum of the scores for each of its categories. 
    \item {\bf Bank Marketing Dataset \cite{Moro2014ADA}:} The protected variable is age (binarized by thresholding at 25).
\end{itemize}

\begin{figure*}[t]
     \centering
     \begin{subfigure}{0.49\linewidth}
     \begin{subfigure}{0.24\linewidth}
        \includegraphics[width=\textwidth]{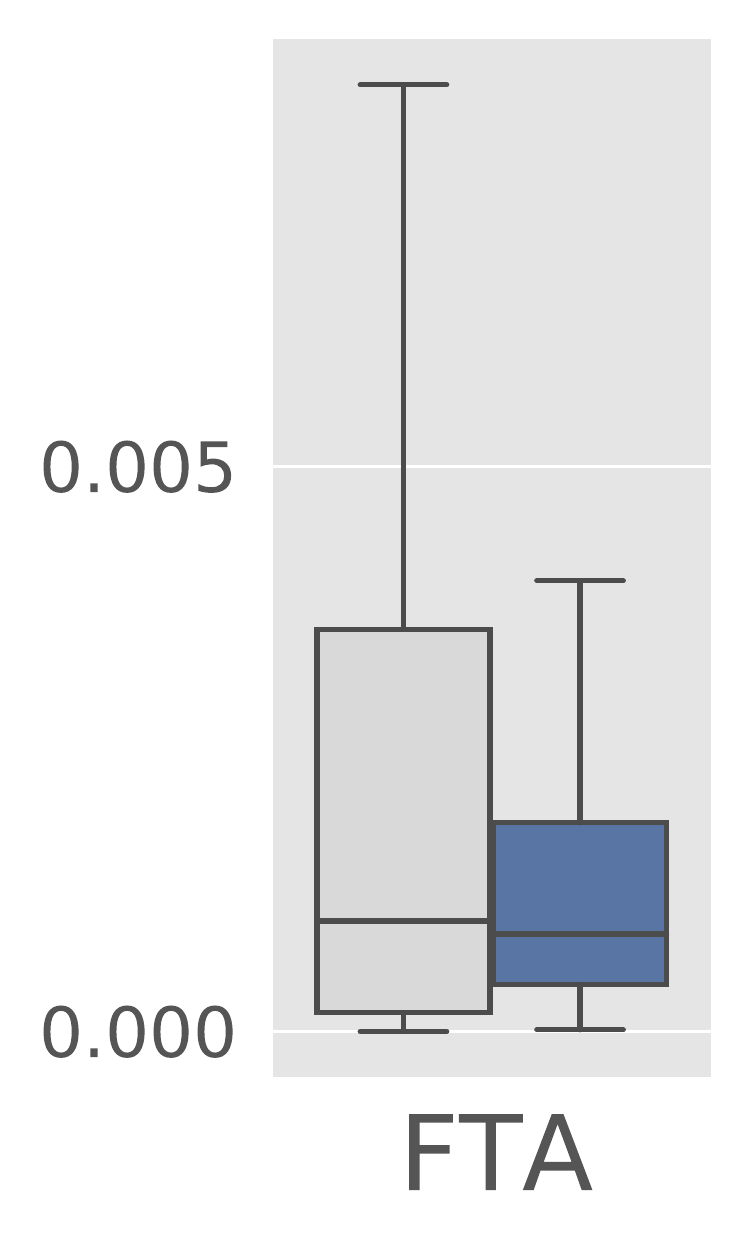}
     \end{subfigure}
     \begin{subfigure}{0.24\linewidth}
        \includegraphics[width=\textwidth]{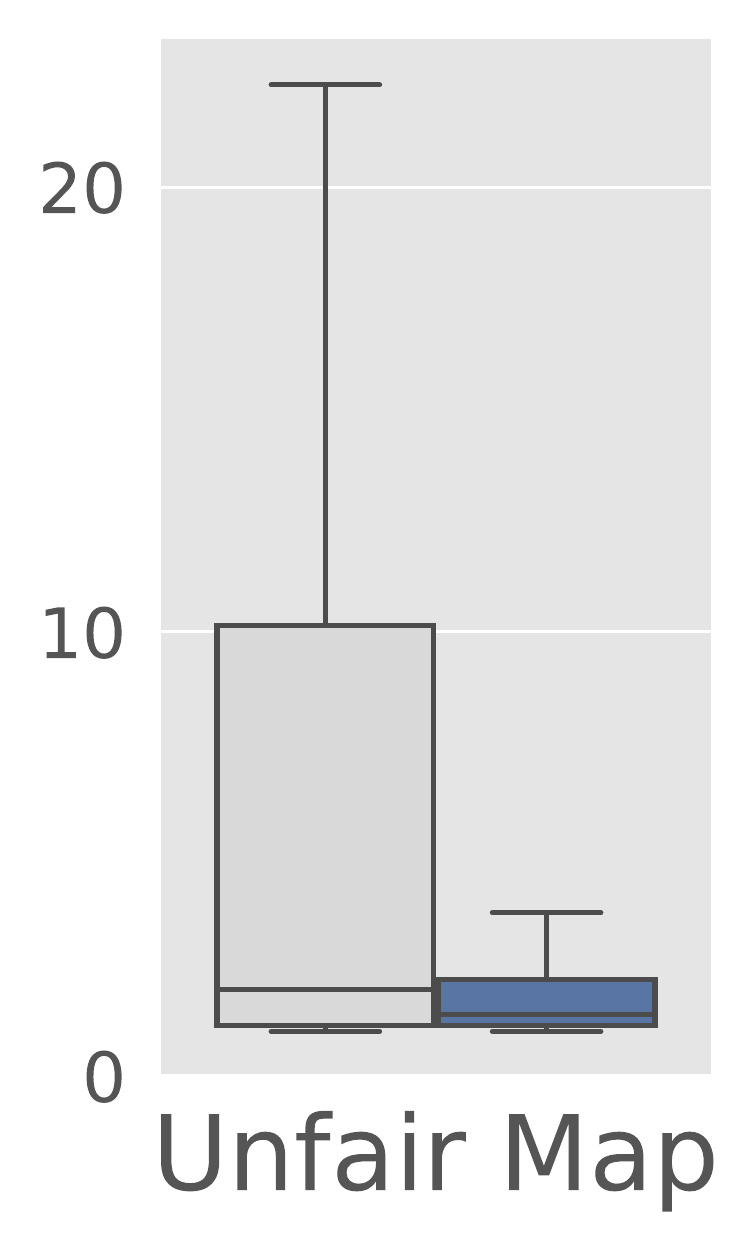}
     \end{subfigure}
     \begin{subfigure}{0.24\linewidth}
        \includegraphics[width=\textwidth]{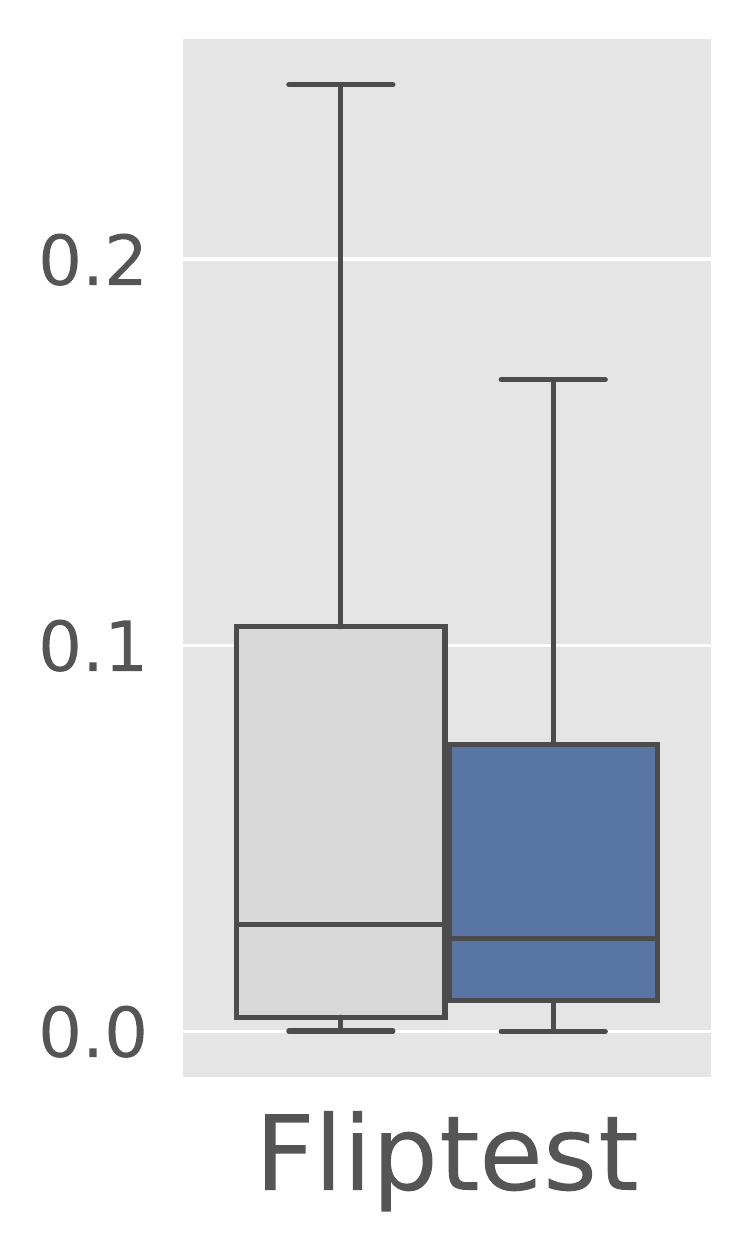}
     \end{subfigure}
     \begin{subfigure}{0.24\linewidth}
        \includegraphics[width=\textwidth]{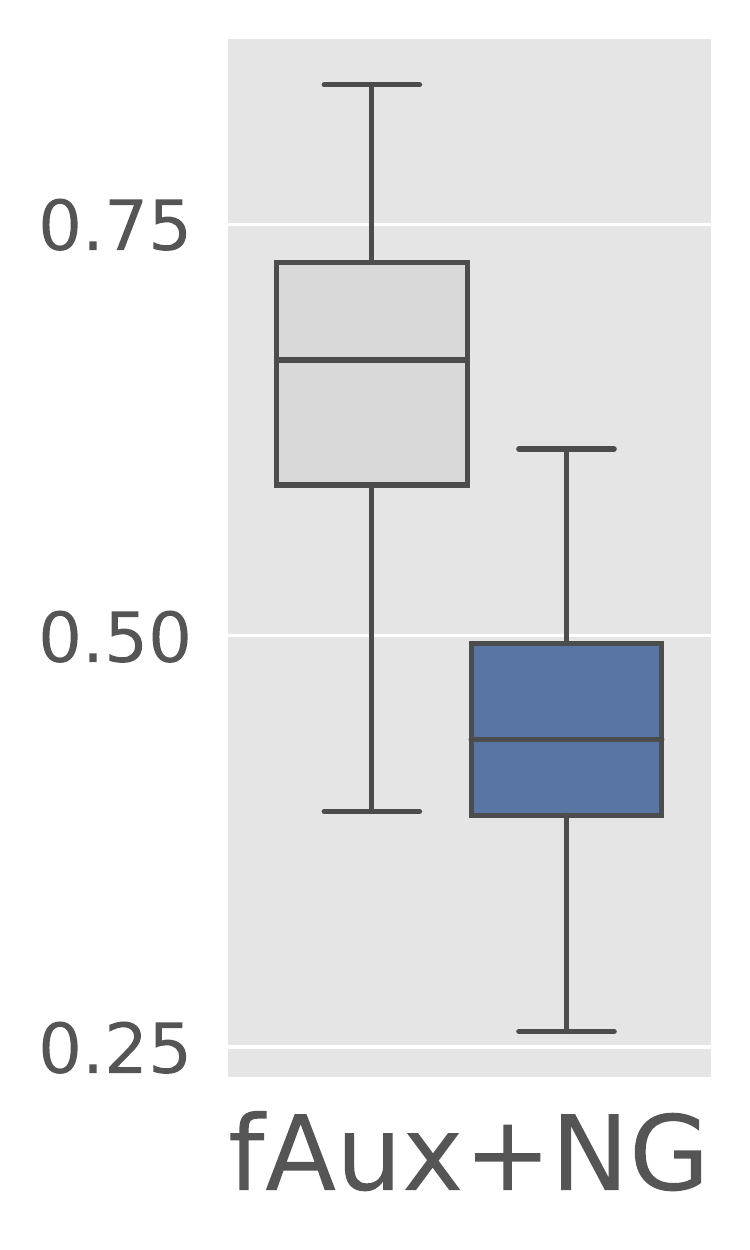}
     \end{subfigure}
     \caption{Adult Dataset}
     \end{subfigure}
     \rulesep
     \begin{subfigure}{0.49\linewidth}
     \begin{subfigure}{0.24\linewidth}
        \includegraphics[width=\textwidth]{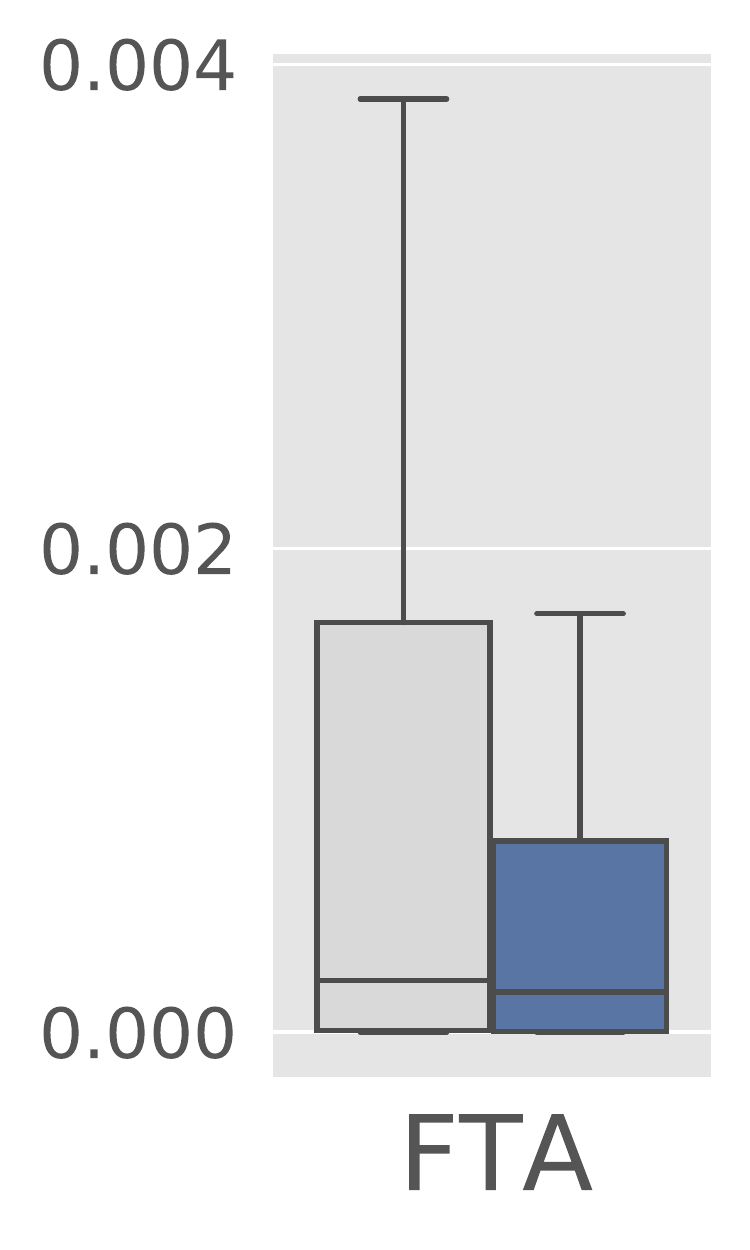}
     \end{subfigure}
     \begin{subfigure}{0.24\linewidth}
        \includegraphics[width=\textwidth]{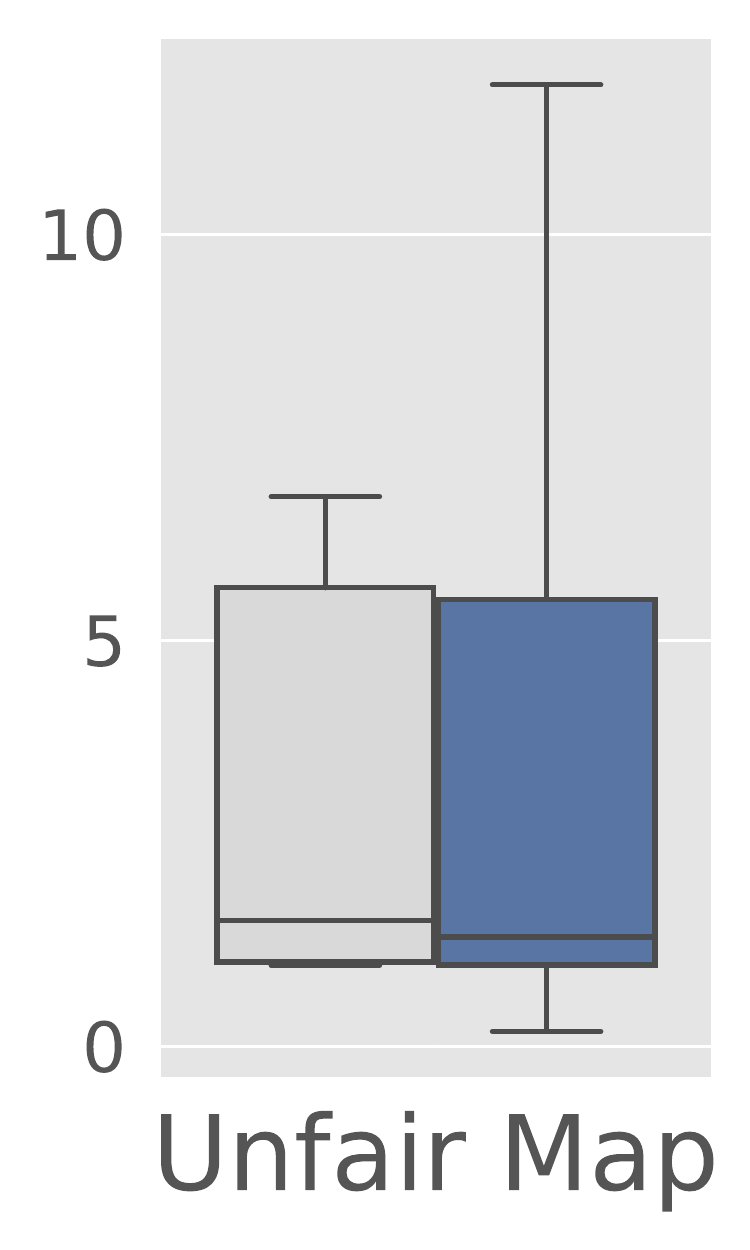}
     \end{subfigure}
     \begin{subfigure}{0.24\linewidth}
        \includegraphics[width=\textwidth]{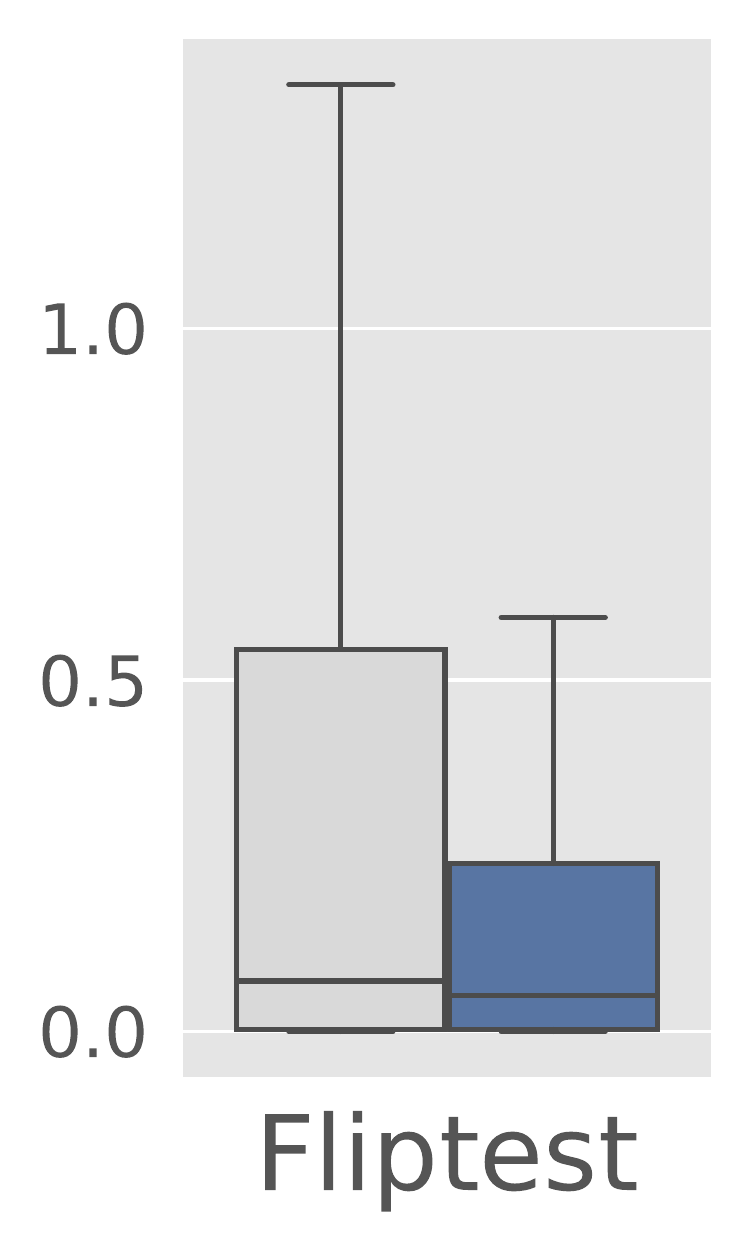}
     \end{subfigure}
     \begin{subfigure}{0.24\linewidth}
        \includegraphics[width=\textwidth]{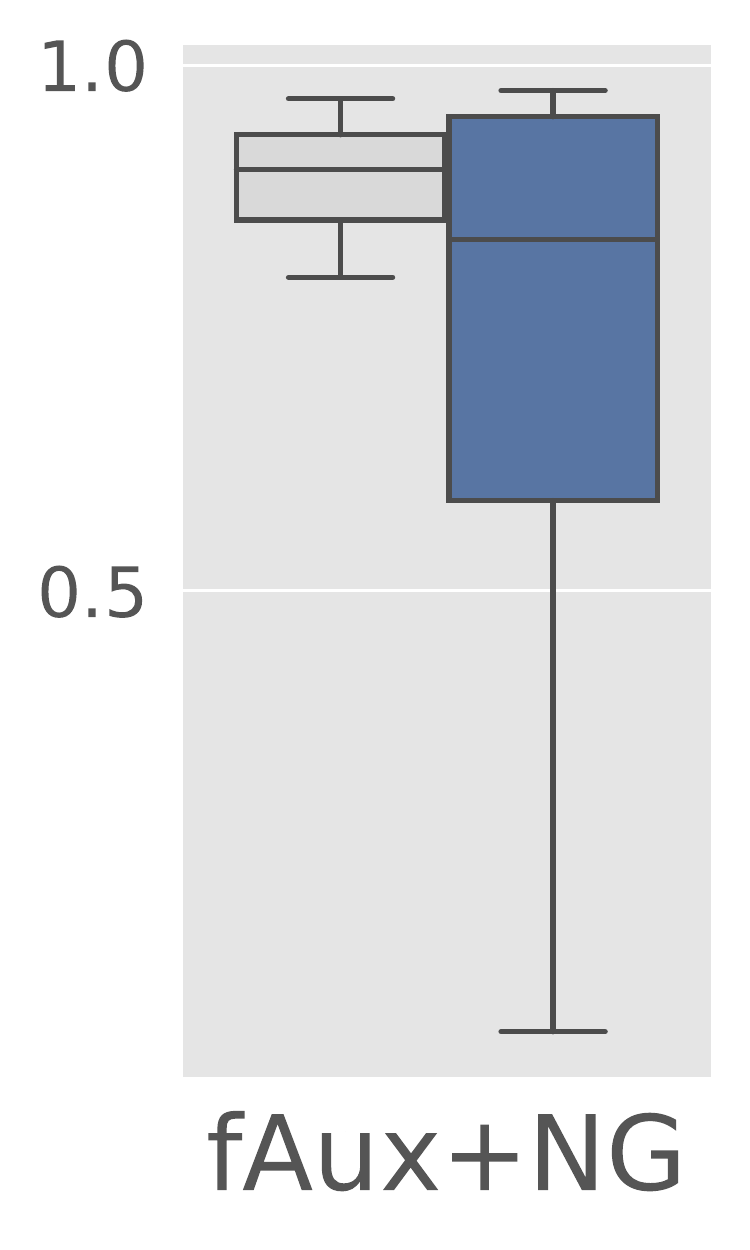}
     \end{subfigure}
     \caption{Bank Marketing Dataset}
     \end{subfigure}
         \caption{\textbf{Performance Comparison among Individual Fairness Testing Methods on Real Datasets.} For each testing algorithm, we show the predicted unfairness score for both unfair model~(grey box) and fair model~(blue box). Greater difference (between the boxes) shows better performance. To conserve space, we have only plotted {\em fAux+NG}.  Plots for the remaining variants are provided in the Appendix, along with additional plots on a third dataset. }
    \label{fig:real_data_exp}
\end{figure*}

\begin{figure*}[t]
     \centering
     \begin{subfigure}{0.32\linewidth}
     \centering
        \includegraphics[width=0.9\textwidth]{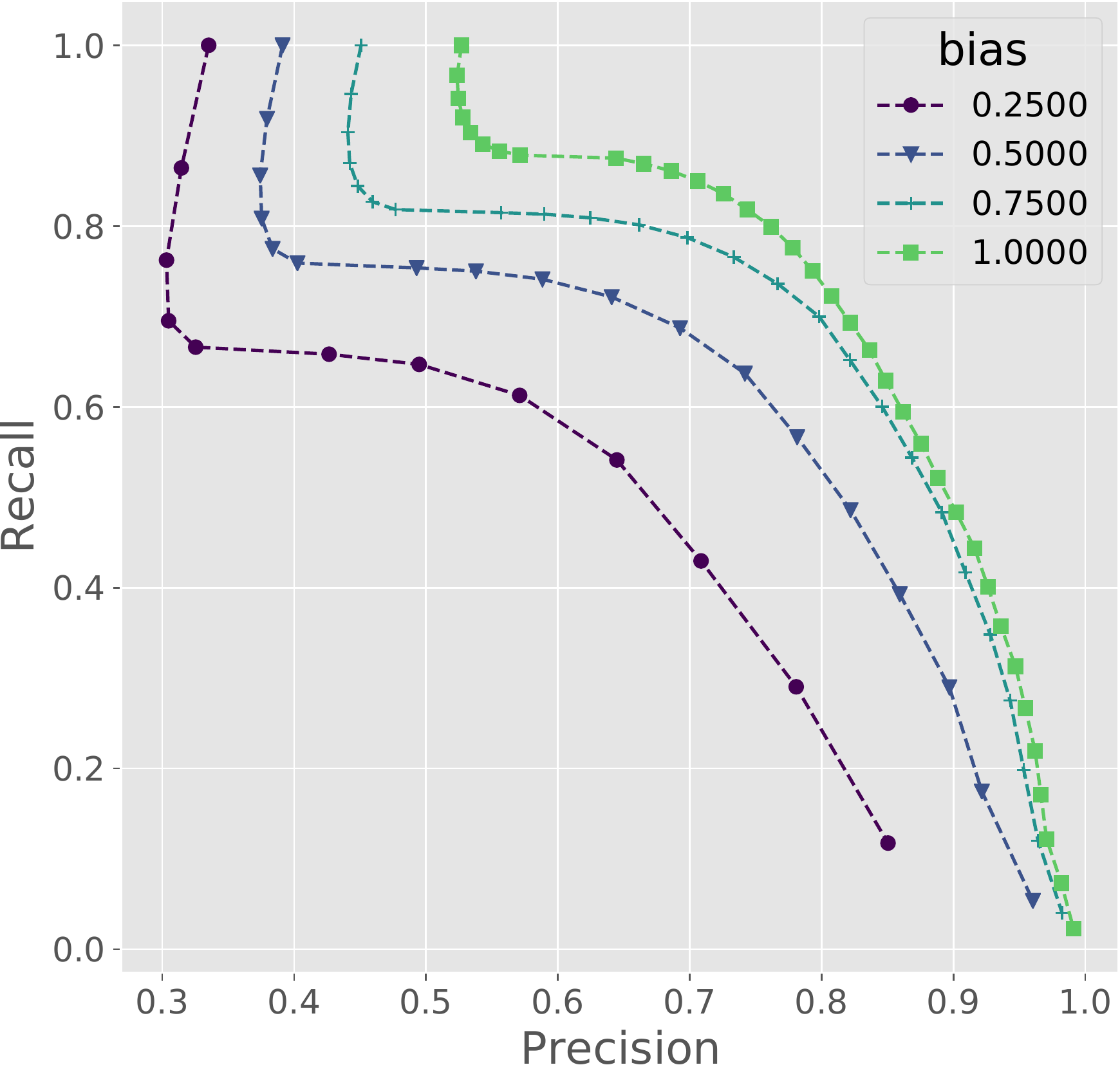}
        \caption{Bias}
        \label{fig:pr:bias}
     \end{subfigure}
     \hspace{\fill}
     \begin{subfigure}{0.32\linewidth}
     \centering
        \includegraphics[width=0.9\textwidth]{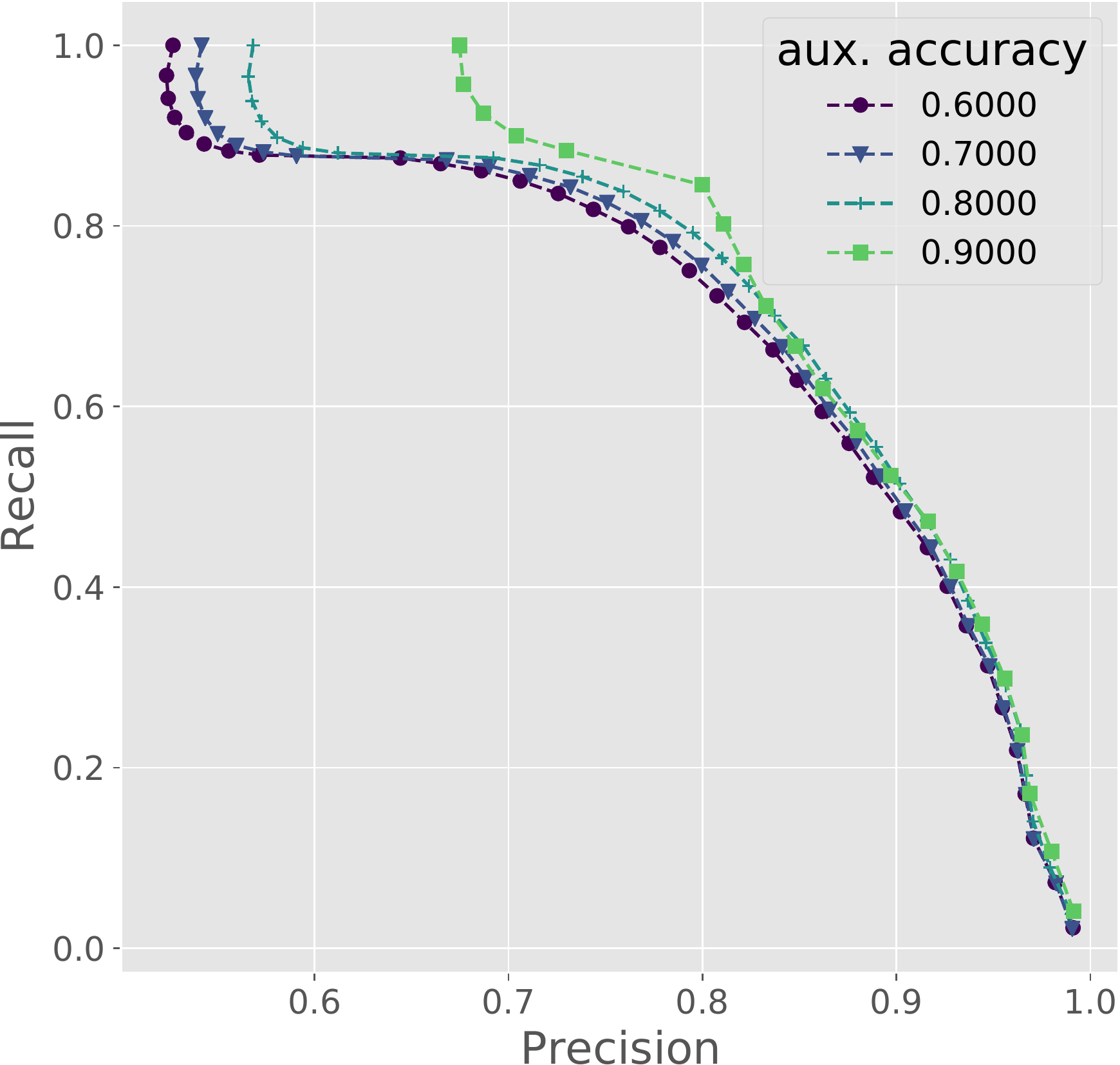}
        \caption{Aux. Model Accuracy}
        \label{fig:pr:acc}
     \end{subfigure}
     \hspace{\fill}
     \begin{subfigure}{0.32\linewidth}
     \centering
        \includegraphics[width=0.9\textwidth]{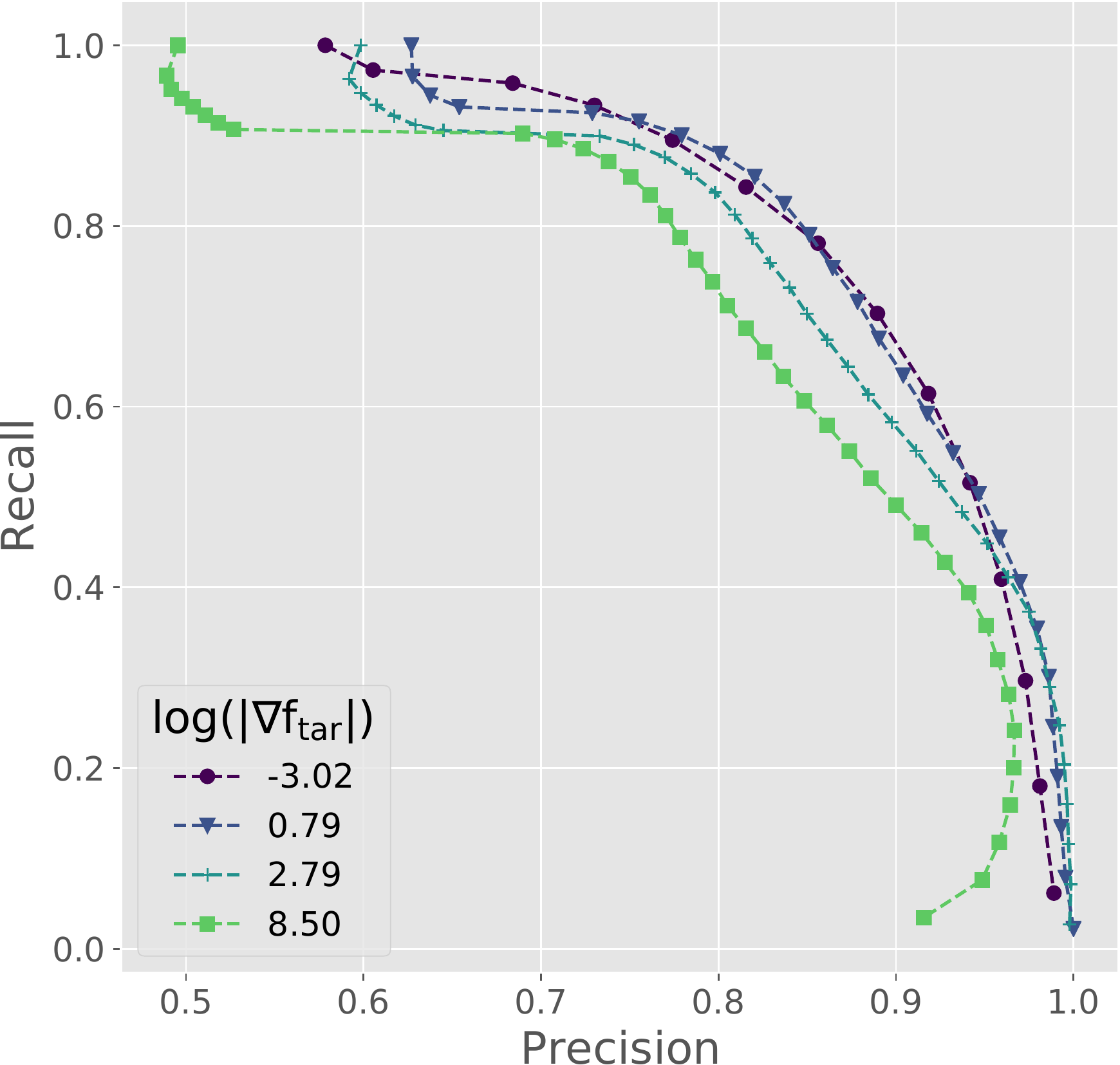}
        \caption{Target Model Gradient}
        \label{fig:pr:ynorm}
     \end{subfigure}
     \hspace{\fill}
    \caption{\textbf{Factors impacting fAux performance.} We identify the operational range of fAux by examining PR curves on aggregated runs.  (a) Impact from data bias. (b) Impact from the auxiliary model performance. (c) Impact from the gradient magnitude of the target model.}
    \label{fig:pr-plots}
\end{figure*}

\subsection{Performance on Synthetic Data}\label{sec:synth-data-exp}

Table~\ref{table:synth-data-res} compares the proposed fAux methods with two state-of-the-art testing approaches on the synthetic datasets. We make the following observations:

\begin{enumerate}[leftmargin=10pt, noitemsep]
    \item We note the fAux variants outperform the weighted $l_p$ norms across all the experiments in the Table~\ref{table:synth-data-res} by a large margin. We attribute this to the more complex relationships captured by our nonlinear auxiliary model.
    %especially when compared on worst-case performance
    \item Among all fAux variants, the fAux+NG shows the most promising test performance.  However, fAux+IG becomes the more reliable test as the accuracy of the auxiliary model decreases. The baseline fAux performs worse than the other variants. This observation reflects our intuition that controlling the variance of gradients is critical to maintaining good performance.
    \item FlipTest demonstrates competitive performance on some datasets but has larger variability in performance than the proposed fAux models. We attribute this variability to the diffuculties of training GANs. In particular, we note the FlipTest performance is much worse on in the outer-product datasets.  These are much higher dimensional, making it more difficult for FlipTest to correctly model the relations through generative models.
\end{enumerate}

\subsection{Performance on Real Data}
In the absence of ground truth labels for discrimination, we cannot quantify the precision of a fairness test at the level of individual datapoints.  However, we may still compare their ability to distinguish fair and unfair models, as is done in \cite{maity2021statistical,yurochkin2021sensei}.  We construct fair models by employing adversarial regularization \cite{OneNetworkFairness}; models become spontaneously unfair in the absence of such regularization. In Figure~\ref{fig:real_data_exp}, we plot the distribution in the test scores for fair and unfair models. For both datasets, we note the fAux approach shows better performance than the existing approaches since it produces observably higher unfairness scores to the Unfair model than those of the Fair model. Moreover, with fAux testing, the difference of score distributions between the Unfair and Fair model is also more distinguishable, which helps the threshold search using statistic tools.

\subsection{Exploring Performance}\label{sec:insights}
While fAux shows good experimental performance, it is important to explore potential conditions that may impact its effectiveness. To analyze the conditions comprehensively, we train target and auxiliary models on the same synthetic datasets previously considered, but over a wider range of settings (Appendix \ref{sec:supp:specific-settings-for-insights}).
We aggregate these results to examine the effectiveness of the fAux test across different datasets.  In particular, we investigated the following aspects:

%The first question is whether the fAux method can correctly identify models that exhibit slight unfair treatment. As the ground truth unfairness level of a target model is hard to measure, we use the bias level as the indicator, as this leads to higher rates of discrimination. 
\begin{itemize}[leftmargin=0pt, label={}]
\item \textbf{Sensitivity:} The first question is whether fAux can correctly identify models that are only slightly discriminatory.  Datasets with a smaller bias parameter (Section \ref{sec:exp:synthetic-exp}) have less correlation between the features and the protected variable, thus we expect that models trained on them will be less discriminatory\footnote{We demonstrate this visually in Figure \ref{fig:cff-bias} of the Appendix.}.  Figure~\ref{fig:pr:bias} shows the Precision-Recall~(PR) curve of the fAux performance based on the degree of the data bias. We note that when the target model is trained on highly biased data ($bias\geq 0.75$), fAux shows better discrimination detection performance. When the target model is trained on less biased data ($bias\leq 0.5$), the reliability of fAux decreases. In the worst-case ($bias \leq 0.25$), fAux loses its functionality since the PR curve and the diagonal line~(the random guess line) have multiple intersections. 
\item \textbf{Auxiliary model accuracy:} Figure~\ref{fig:pr:acc} shows the PR curve for different auxiliary model accuracies. 
Model accuracy varies according to the hyperparameters used, and the difficulty of the classification task.  Empirically, we find that auxiliary models with better accuracy lead to better fAux performance. Since the auxiliary models aim to predict the protected variables $C$, their performance impacts how well fAux detects the variance of $C$ given features $X$. %However, we also note that the fAux methods are not sensitive to the auxiliary model's accuracy as all of the PR curves have large overlaps.
%%%\Simon{ This used to say target model accuracy, which I assume was a mistake.}
%%%\Simon{ Need to add something about how you manipulated the accuracy of the auxiliary model.}
% this figure shows the accuracy of auxiliary model not target model.
\item\textbf{Quality of target model:} In practice, target model quality is out of scope for tuning fairness testing methods. However, we note that the gradient magnitude $|\nabla f_{tar}|$ impacts the fAux performance as shown in Figure~\ref{fig:pr:ynorm}. When the gradient is extremely large, fAux exhibits worse performance.
\end{itemize}

% Hmmm.. how confident is confident...
% In order to be confident with the predictions of fAux, it is important to understand under which conditions the approximation breaks down.  To this end, we repeat the experiments of Section \ref{sec:benchmarks} across a wider series of datasets and biases.  We then aggregate the results of the experiments together to examine the precision and recall across different models and datasets.  More detailed descriptions of the results and experimental configurations can be found in the supplementary materials.  We summarize some of the larger sources of variation in Figure \ref{fig:pr-plots}.  The most critical of these is the amount of bias in the dataset (Figure \ref{fig:pr:bias}), as lower amounts make it difficult to distinguish discrimination from other modelling errors.  Following this,  auxiliary models which more accurately predict $C$ (Figure \ref{fig:pr:acc}) are better at detecting changes in $C$. 
% I have trouble to parse the following sentence
%Finally, many of the normalization schemes used in fAux break down when the norm of the target model gradient $|\nabla f_{tar}|_2$ is itself large (Figure \ref{fig:pr:ynorm}), and we recommend abstaining from flagging discrimination in these cases.

\section{Conclusion}
In this paper, we proposed a novel criterion for testing individual fairness (Theorem \ref{theorem:LIC}), which we related to other well-known definitions (summarized in Definition \ref{def:canonical}).  This new criterion is based on locally testing model derivatives, and thus avoids the challenge of generating similar pairs of inputs.  In our experiments, we have demonstrated that our new test - named fAux - outperforms several other state-of-the-art approaches on both synthetic and real datasets. 

%It has the further advantages that it does not require manually defined distance metrics since it uses derivatives to infer local discrepancies.  Moreover, it does not require a generative model to support testing and so there is no need to validating the quality of the generated counterexamples.  

%However, in this paper, we have limited our focus to datasets with historical bias, and more work must be done to adapt fAux to accommodate other sources.  Finally, because of our emphasis on local methods and simple models, future work may investigate building formal tools to provide certificates for individual fairness.

%%%\Simon{We should add something about (i) Limitations of the model and (ii) possible extensions here.  Ideas?}
% yes, I cut it out as below to reduce the space before. But now we have space for this. :P

% In the experiments, we show the fAux outperforms several state-of-the-art verification approaches on both synthetic datasets and real datasets. %Finally, we investigated the conditions that keep fAux verification reliable.

\bibliographystyle{named}
\bibliography{references}

\clearpage

\appendix

\section{Appendix}
\subsection{Recap the Practical Usage of the Proposed Testing Approach}\label{sec:recap}
%\chris{some of this content should be added to the main paper to help with intuition}

In order to introduce individual fairness testing into industry as an important checking point/regulation, an efficient and effective testing approach that can scale up to applications with millions of customers is urgently needed. We note that the existing approaches such as counterfactual fairness, while conceptionally effective, are hard to deploy in practice as they either consume too much computation resources (that are out of budget) or require detailed domain knowledge (that are usually too complex to summarize). Hence, the proposed approach becomes the most practical option. It only needs to estimate gradient (or feature importance) alignment, which happens to also be unavoidable when testing for other properties such as adversarial robustness testing or performing feature importance analysis, etc.

That said, we caution against over reliance on mathematical metrics for quantifying fairness.  In particular, while we show in Table \ref{table:comp-eff} that faux can flag proxy features, we emphasize this is no substitute for intuition about the impact of the features in the dataset.  We advocate faux as an effective component of a validation pipeline, but investigators should be mindful of other sources of bias beyond historical bias e.g. label contamination.

\subsection{Use Cases}\label{sec:use-cases}

\subsubsection{Predictive Policing}

%\chris{this is bold and potentially controversial so make sure everything is cited properly and that you don't inject any opinions into the writing/need to be factual and objective.}

We consider the use case described in \cite{CounterfactualFairness}, where a hypothetical city government wants to predict crime rates in order to assign policing resources to different communities.  The goal is that, given an individual, a model may be used to estimate that individual's predisposition to violence.  A developer trains a model on a dataset obtained by merging residential information with police records of arrests.  However, some neighborhoods have higher arrest rates due to greater policing there.  Because individuals of different races may congregate in different neighborhoods, this can lead the model to conclude that members of a particular race are more likely to break the law.  

This bias persists even if the classifier does not take race as an explicit input, since that information may be inferred from the neighborhood.  Thus, Fairness Through Unawareness (FTU) is not applicable. fAux, however, can identify such correlations through the gradient of the auxiliary model.  Moreover, it can recognize when these correlations impact model predictions through gradient alignment.

\subsubsection{Credit Cards}

Consider a hypothetical credit agency that wants to automate the process of giving credit cards.  As inputs, they merge account data, transaction history, and demographic information to form a high-dimensional, sparse dataset with millions of entries.  They train a model to predict whether the individual will default on their payments within a certain time window.  The model learns that individuals above a certain income threshold are more likely to make their payments.  However, due to income disparity between men and women, and income disparity between younger and older individuals, these individuals may be unfairly treated by the model.

Fairness Through Awareness (FTA) may be deployed to solve this problem, by going through pairs of individuals, and determining if similar treatment is offered to similar individuals (as measured by a similarity metric).  However, this has a few shortcomings.  Firstly, computing pairwise similarities is computationally daunting, especially in higher dimensions.  Secondly, the dataset is constructed from a mixture of tabular and time series data, meaning it incorporates different data domains that have different semantic interpretation.  Finally, the model may be justified in treating two "similar" individuals differently if they differ in a business-relevant sensitive feature.  Depending on the similarity metric, FTA may completely miss the underlying issue, which in this case, is the income disparity, and simply identify all instances where the model was sensitive.

By contrast, fAux avoids the need for designing similarity metrics, by inferring which features are relevant for predicting sex/age.  By aligning the feature importances with the the gradients of the target model, fAux can identify when the model is leveraging this information to make its predictions, revealing the discrimination.

% be cautious of using this example. It is a sensitive topic nowadays.
\subsection{Discussion: Relation to Adversarial Robustness}
\label{sec:discussion_relation_ar}
Individual fairness testing has strong connections with adversarial robustness~\cite{yurochkin2020training,john2020verifying}. The adversarial robustness criterion near point $\mathbf{x}$
\begin{equation}
    \forall \mathbf{x}^{\prime} \,s.t  \,  \left(\phi_{in}(\mathbf{x}, \mathbf{x}^{\prime})  \leq \epsilon \right) \land 
    \left(\phi_{out}(f(\mathbf{x}), f(\mathbf{x}^{\prime}))  < \delta \right)
    \label{eq:adversarial_criterion}
\end{equation}
is nearly identical to the criterion of the FTA definition if we reformulate the FTA criterion as
\begin{equation}
    \forall \mathbf{x}_i \mathbf{x}_j \,s.t  \,  \left(\phi_{in}(\mathbf{x}_i, \mathbf{x}_j)  \leq \epsilon \right) \land 
    \left(\phi_{out}(f_{tar}(\mathbf{x}_i), f_{tar}(\mathbf{x}_j))  < \delta \right)
    \label{eq:adversarial_criterion}
\end{equation}
with additional constraint $\epsilon \geq \delta$. Similarly, the counterfactual fairness criterion, can be rewritten as
\begin{equation}
\begin{aligned}
    &\forall \mathbf{c}' \,s.t \, \\
    &\left(\phi_{in}(\mathbf{c}, \mathbf{c}')\!\leq\!\epsilon \right) \!\land\!
    \left(\phi_{out}(P(y|\mathbf{x},\mathbf{c})\!-\!P_{C\leftarrow\textit{do}(\mathbf{c}')}(y| \mathbf{x}, \mathbf{c}))) \!<\!\delta \right)
\end{aligned}
    \label{eq:adversarial_criterion}
\end{equation}
to match the adversarial robustness definition. 
%\Christopher{See my previous comment in section 2.2.3, do we need the "do" operation?} 
% wuga: yes, we do. lol

The Local Fairness Test proposed in this paper also has strong connections with adversarial robustness. However, as fAux works on the level of differentiation, the input distance condition $\phi_{in}(\mathbf{c}, \mathbf{c}+\Delta\mathbf{c})$ becomes trivial since $\Delta\mathbf{c}$ is negligible.
%\Christopher{This is not clear to me. Can you expand on why this is the case?}
Hence, fAux considers only the output distance condition:
% \begin{equation}
%     \phi_{out} = \left|\frac{\nabla f_{tar}^{\top}\nabla f_{aux}}{\nabla f_{aux}^{\top}\nabla f_{aux}}\right|,
% \end{equation}
\begin{equation}
    \phi_{out} = \left|\nabla f_{tar} \left(\nabla f_{aux}^\top\nabla f_{aux}\right)^{-1}\nabla f_{aux}^\top \right|_{\infty},
\end{equation}
where distance function over the inputs is no longer explicitly provided; the derivative is the limit of the predictive difference between two near identical inputs. 
%\Christopher{I;m having trouble to understand to last sentence? Can you rephrase?}
% wuga: changed to derivative.. I think the differentiation caused confusion

However, even though individual fairness testing and adversarial robustness testing share common properties mentioned above, we note that they are different tests that aim to reveal different weaknesses of a machine learning model. Specifically, adversarial robustness testing focuses on detecting vulnerability of a model with respect to anomalous (or adversarial) inputs, whereas individual fairness testing pays attention to the sensitivity of protected attributes on a model.
\subsection{Proof of Theorem 3.1}
\label{sec:theorem_proof}
The $l_\infty$ norm represents a max operation on the most violated protected attribute c in $\mathbf{c}=\{c_1,c_2\cdots c_k\}$. To prove theorem 3.1, we only need to focus on this single protected attribute $c$. Hence, if a prediction violates the \textit{Local Independence Criterion (LIC)}, we can express it as:
\begin{equation}
\begin{aligned}
%\max_{k}
\left|\frac{\partial f_{tar}( \mathbf{x})}{\partial c}\right|
    %&= \lim_{\Delta{\mathbf{c}}\to 0} \frac{f_{tar}(f_g(\mathbf{z}_{\bot}, \psi(\mathbf{c}))) - f_{tar}(f_g(\mathbf{z}_{\bot}, \psi(\mathbf{c}+\Delta\mathbf{c})))}{\Delta{\mathbf{c}}}\\
    %\leq \lim_{\Delta{\mathbf{c}}\to 0}\frac{\delta}{\Delta{\mathbf{c}}},
    > \delta
\end{aligned}
\label{eq:local_independent_definition}
\end{equation}

By expanding the left hand side of the inequality with the generative model $f_g$, we note
\begin{equation}
\begin{aligned}
%\max_{k}
% \left|\frac{\partial f_{tar}( \mathbf{x})}{\partial c}\right|
    \lim_{\Delta{c}\to 0} \frac{f_{tar}(f_g(\mathbf{z}_{\bot}, \psi(c))) - f_{tar}(f_g(\mathbf{z}_{\bot}, \psi(c+\Delta c)))}{\Delta{c}}
    %\leq \lim_{\Delta{\mathbf{c}}\to 0}\frac{\delta}{\Delta{\mathbf{c}}},
    %\leq \delta^{\prime}
\end{aligned}
\label{eq:expand_with_limit}
\end{equation}
which could be numerically approximated with small $\Delta c$ such that
\begin{equation}
\begin{aligned}
%\max_{k}
% \left|\frac{\partial f_{tar}( \mathbf{x})}{\partial c}\right|
    &\lim_{\Delta{c}\to 0} \frac{f_{tar}(f_g(\mathbf{z}_{\bot}, \psi(c))) - f_{tar}(f_g(\mathbf{z}_{\bot}, \psi(c+\Delta c)))}{\Delta{c}}\\
    &= \frac{f_{tar}(f_g(\mathbf{z}_{\bot}, \psi(c))) - f_{tar}(f_g(\mathbf{z}_{\bot}, \psi(c+\Delta c)))}{\Delta c} + \xi,
    %\leq \lim_{\Delta{\mathbf{c}}\to 0}\frac{\delta}{\Delta{\mathbf{c}}},
    %\leq \delta^{\prime}
\end{aligned}
\label{eq:expand_with_approximation}
\end{equation}
where the $\xi$ is an error term introduced by the approximation.

Therefore, Inequality~\ref{eq:local_independent_definition} could be extended as
\begin{equation}
\begin{aligned}
&\left|f_{tar}(f_g(\mathbf{z}_{\bot}, \psi(c))) - f_{tar}(f_g(\mathbf{z}_{\bot}, \psi(c+\Delta c)))\right| \\
&> (\delta-\xi) |\Delta c|.\\
\end{aligned}
\label{eq:local_independent_definition_extended}
\end{equation}
If we choose $\delta \gg\xi$ and $|\Delta c|$ is positive, we have
\begin{equation}
\begin{aligned}
&\left|f_{tar}(f_g(\mathbf{z}_{\bot}, \psi(c))) - f_{tar}(f_g(\mathbf{z}_{\bot}, \psi(c+\Delta c)))\right| > 0,\\
\end{aligned}
\end{equation}
which violates the individual fairness definition in Definition~\ref{def:canonical}.

% {\color{red} To be clear, the above proof only shows that there must exist a $\delta'$ such that the global flipping threshold $\delta$ could be satisfied. In this work, we treat $\delta'$ as an independent hyper-parameter (or threshold setting) that not necessarily related to $|\Delta c|$. In fact, the threshold of $\delta'$ should be tuned on validation dataset to maximize the testing accuracy.}
\subsection{Using Pseudoinverses for Model Inversion}\label{sec:pseudo}

In this section, we provide a more in-depth discussion on the derivation of Equation 11.  The Taylor expansion of Equation 10 is a local linear approximation of the function $f_{aux}$. The left-hand side of this equation denotes the change in the score, and is a scalar.  The right hand side is a dot product between two vectors.  While we would like to invert this equation to solve for $\textbf{x}$, Equation 11 is an indeterminate equation .  Nevertheless, using the Moore-Penrose pseudoinverse, we may still characterize the infinite solution set for $x$.

To understand this, consider a general linear equation:
\begin{equation}
    \textbf{y} = A\textbf{x}. \label{eq:gen-linear}
\end{equation}
Here, $A$ is an $m$ by $n$ matrix, and we want to solve this equation for $\textbf{x}$.  When $A$ does not have an inverse, solutions may be found to \eqref{eq:gen-linear}, but they will not be unique.  Instead, there will be a family of solutions $x(z)$ parameterized by a vector $z \in \mathbf{R}^{n}$, which will be given by:
\begin{equation}
    \textbf{x}(\textbf{z}) = A^{\dagger} \textbf{y} + (I - A^{\dagger} A) \textbf{z}.\label{eq:linear-sol}
\end{equation}
Here, $I$ is the $n$ by $n$ identity matrix, and the pseudo-inverse is given by %\Christopher{may want to add a citation}:
\begin{equation}
    A^{\dagger} = A^{\top}(A A^{\top})^{-1}.\label{eq:Adagger}
\end{equation}
Note that equation \ref{eq:linear-sol} describes an infinite solution set, insofar as $\textbf{z}$ is undetermined: the operation of $A$ on the right-hand side of \eqref{eq:linear-sol} will simply cause the second term to vanish, so that any vector $z$ may be used.  

For the specific case of our local linear approximation of the auxiliary model, $y = (c - f_{aux}(\textbf{x}_0)$ is a scalar, $A = \nabla f_{aux}^{\top}$ is a co-vector, and we may solve for $f_{sur}(c)$ as:

\begin{equation}
\begin{aligned}
    &f_{sur}(c) =\\
    &\textbf{x}_0\!+\!(c\!-\!f(\textbf{x}_0))\frac{\nabla f_{aux}}{\nabla f_{aux}^{\top}\nabla f_{aux}}\!+\!(I\!-\!\frac{\nabla f_{aux}\nabla f_{aux}^{\top}}{\nabla f_{aux}^{\top}\nabla f_{aux}}) \textbf{z} 
\end{aligned}\label{eq:penrose-invers}
\end{equation}

\begin{figure}[t!]
\centering
\includegraphics[trim={0 50 0 0},clip, width=0.4\textwidth]{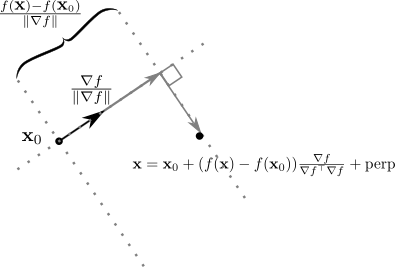}
\caption{\textbf{Inverting a linear function} In one dimension, the inverse of a linear function is uniquely determined by the slope.  In higher dimensions, the slope becomes the magnitude of the gradient.  However the inverse is no longer uniquely determined: the output is constant along directions perpendicular to the gradient.}
\label{fig:penrose-illustration}
\end{figure}

\noindent We illustrate these different terms in Figure \ref{fig:penrose-illustration}. Intuitively, the equation above describes two sources of variation: the first of which is parallel to $\nabla f_{aux}$, and the second of which is perpendicular to it.  The partial derivative with respect to $C$ is defined by the variance in a function as $C$ is changed, and all other variables are fixed.  As a relaxation, we consider the direction that leads to maximal change in $C$, and minimal change in all other variables.  This corresponds to the minimum norm solution that satisfies \eqref{eq:penrose-invers}.  

For a model trained to predict $c$, the gradient should point in the direction of greatest change in $c$.  Therefore, intuitively, the $c$-dependence of the perpendicular part should be much smaller than the $c$-dependence of the parallel part.  We may thus disregard the perpendicular part, and are left with the following approximation:

\begin{equation}
    \frac{\partial \mathbf{x}}{\partial \mathbf{c}} \approx  \frac{\nabla f_{aux}}{\nabla f_{aux}^{\top}\nabla f_{aux}}\label{eq:partial-x-approx}
\end{equation}

\noindent

\subsection{Algorithm}\label{sec:aux-model-alg}
Algorithm~\ref{alg:local-independence-test} summarizes the proposed fairness testing approach~(fAux) described in the main paper.

\begin{algorithm}[h]
\SetAlgoLined
\KwResult{Flag unfair model behaviour}
\KwInput{Validation data points $D = \{\cdots(\mathbf{x}_i, \mathbf{c}_i, y_i)\cdots\}$, target model $f_{tar}$, and threshold $\delta'$}
Train auxiliary model $\mathbf{c}=f_{aux}(\mathbf{x})$;

\For{each data point $(\mathbf{x}, \mathbf{c}, y)$ in $D$}{
    Evaluate gradients $\nabla f_{tar}$ and $\nabla f_{aux}$;
    
    Flag unfair behaviour on inputs through %$\left|\frac{\nabla f_{tar}^{\top}\nabla f_{aux}}{\nabla f_{aux}^{\top}\nabla f_{aux}}\right| > \delta' $;
    $\left|\nabla f_{tar} \left(\nabla f_{aux}^\top\nabla f_{aux}\right)^{-1}\nabla f_{aux}^\top \right|_{\infty} \leq \delta'$;
    }

 \caption{Auxiliary Model Test (fAux)}
 \label{alg:local-independence-test}
\end{algorithm}
\subsection{Description of the Synthetic Dataset Pipeline}
\label{sec:synthetic_data_generation}
In Section 2.2 we hypothesized that when datasets contain multiple sources of variation, models will preferentially exploit the simplest patterns to achieve their performance objective.  In this section, we describe a simple framework for constructing synthetic datasets that trigger this misbehaviour.  Specifically, we present a scheme for fusing real datasets together, so that the final dataset contains multiple patterns of variation.  This has the advantage that the final dataset inherits the same types of noisy and nonlinear relationships that are present in real datasets.  We summarize this procedure in Algorithm  \ref{alg:synthetic-dataset}.

The framework has three main degrees of freedom:

\begin{enumerate}[leftmargin=10pt, itemsep=0.1em, topsep=0pt]
    \item To simulate a historical bias\footnote{In this paper we focus on historical biases, though the proposed framework is sufficiently general to handle other mechanisms of bias.}, we allow the target variable $y$ and the protected variable $c$ to be correlated while sampling from the dataset. This is controlled by a joint distribution $P(C, Y)$.
    \item The fusion function $F$, which combines the datasets together without distorting the underlying patterns.  We consider two extremes: one in which data is perfectly mixed (outer product), and one in which it is trivially separable (concatenation).
    \item The choice of datasets, which controls the difficulty of the learning task.  We measure difficulty according to the best classification score that is obtained on each individual dataset using an architecture search.  We build \textbf{Conditional Variational Autoencoders} to model these datasets.
\end{enumerate}

In the next few subsections, we provide more details about each of these components

\begin{figure}[t]
     \centering
     \begin{subfigure}[b]{0.48\linewidth}
        \centering
        \includegraphics[width=\textwidth]{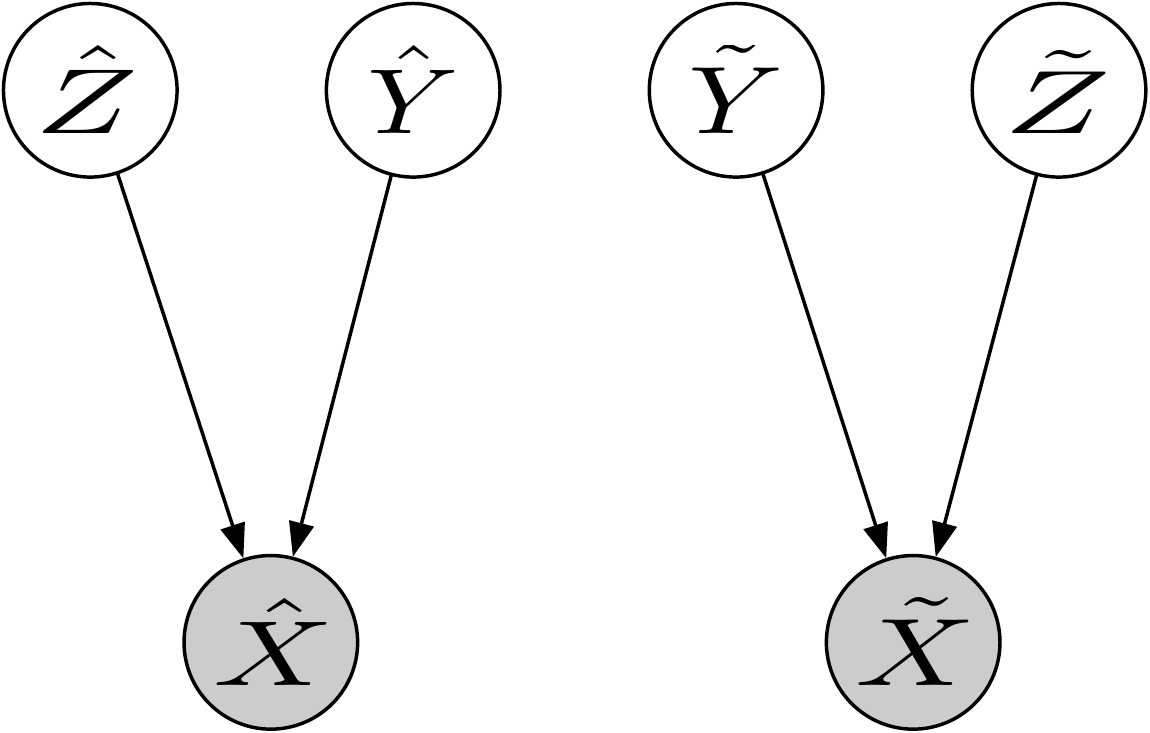}
        % \begin{tikzpicture}
        %     % nodes
        %     \node[latent] (z) {$Z_1$};%
        %     %\node[latent,below=of x,xshift=-1cm,fill] (z) {$z$}; %
        %     \node[latent,below=of z,xshift=0cm] (y) {$Y_1$}; %
            
        %     %\node[latent,below=of x,xshift=-1cm,fill] (z) {$z$}; %
        %     \node[obs,right=of y,xshift=0cm] (x) {$X_1$};
        %     %\draw (0,1.69) circle(.36cm);
        %     % plate
        %     %\plate [inner sep=.25cm,yshift=.2cm] {plate1} {(x)(y)(z)} {$N$}; %
        %     % edges
        %     \edge {y,z} {x} 
        %     %---------
        %     %---------
        %     %---------

        %     %\node[latent,below=of x,xshift=-1cm,fill] (z) {$z$}; %
        %     \node[latent, right=of x,xshift=0cm] (y2) {$Y_2$}; %
            
        %     \node[latent, above=of y2, xshift=0cm] (z2) {$Z_2$};%
            
        %     %\node[latent,below=of x,xshift=-1cm,fill] (z) {$z$}; %
        %     \node[obs,right=of y2,xshift=0cm] (x2) {$X_2$};
        %     %\draw (0,1.69) circle(.36cm);
        %     % plate
        %     %\plate [inner sep=.25cm,yshift=.2cm] {plate1} {(x)(y)(z)} {$N$}; %
        %     % edges
        %     \edge {y2,z2} {x2} 
        % \end{tikzpicture}

         \caption{Unbiased datasets}
         \label{fig:synthetic-bias:no-bias}
     \end{subfigure}
     \hspace{\fill}
     \begin{subfigure}[b]{0.48\linewidth}
        \centering
        \includegraphics[width=\textwidth]{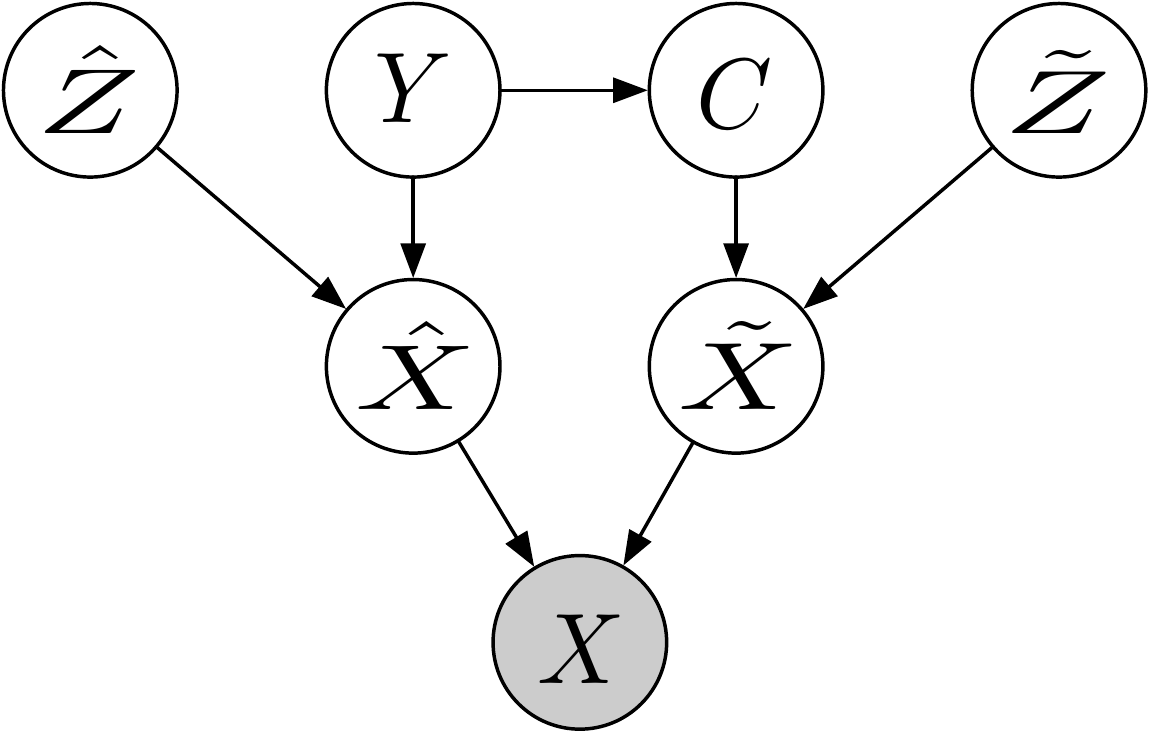}
        % \begin{tikzpicture}
        %     % nodes
        %     \node[latent] (z1) {$Z_1$};%
        %     %\node[latent,below=of x,xshift=-1cm,fill] (z) {$z$}; %
        %     \node[latent,below=of z,xshift=0cm] (y) {$Y_1$}; %
            
        %     %\node[latent,below=of x,xshift=-1cm,fill] (z) {$z$}; %
        %     \node[latent,right=of y,xshift=0cm] (x1) {$X_1$};

        %     \edge {y,z1} {x1}
            
        %     \node[latent,below=of y,xshift=0cm] (c) {$C$}; %
    
        %     \node[latent,below=of c,xshift=0cm] (z2) {$Z_2$}; %
            
        %     \node[latent,right=of c,xshift=0cm] (x2) {$X_2$};
            
        %     \edge {y}{c}
            
        %     \edge {c,z2} {x2} 
            
        %     \node[obs,right=of x1,yshift=-0.75cm] (x) {$X$};
            
        %     \edge {x1,x2}{x}
        % \end{tikzpicture}

         \caption{Fusing synthetic bias}
         \label{fig:synthetic-bias:bias}
     \end{subfigure}
        \caption{\textbf{The synthetic bias pipeline.} (a) Generative models for unbiased datasets.  In both datasets the observable features $X$ are distributed according to a target $Y$ and latent variables $Z$.  At present there are no distinct populations in either dataset, and so there is no bias.  (b) $X_1$ and $X_2$ are fused together to form a new dataset $X$, which contains two distinct patterns of variation.  We declare the target of the second dataset $Y_2$ to be a sensitive attribute, labeled with $C$.  We introduce a historical bias using a correlation between $Y$ and $C$.  Models trained on the new dataset may learn to exploit the $C$ dependence of $X$ to predict $Y$, which corresponds to discriminatory behaviours.}
        \label{fig:synthetic-bias}
\end{figure}

\begin{algorithm}[t]
\SetAlgoLined
%\SetKwInOut{Output}{Output}
\SetKwInOut{Input}{Input}
%\SetKwInOut{Params}{Params}

\KwResult{Obtain samples $\textbf{x}$ from a biased dataset}
\Input{Joint distribution $Pr(c,y)$, latent priors $Pr(\hat{z})$, $Pr(\tilde{z})$, generative models $\Gamma_y$, $\Gamma_c$, fusion function $F$}
%\KwData{}

\begin{enumerate}
    \item Sample $\hat{z} \sim Pr(\hat{z})$ and  $\tilde{z} \sim Pr(\tilde{z})$.
    \item Sample $(y,c) \sim Pr(c,y)$
    \item Compute $\hat{\mathbf{x}}_i = \Gamma_y(\hat{z}, y)$, $\tilde{\mathbf{x}}_j = \Gamma_c(\tilde{z}, c)$
    \item Compute $\textbf{x} = F(\hat{\mathbf{x}}_i, \tilde{\mathbf{x}}_j)$
\end{enumerate}

 \caption{Synthetic Data Generation}
 \label{alg:synthetic-dataset}
\end{algorithm}

\subsubsection{Bias parameter}\label{sec:supp:bias-param}

We can control the amount of bias by specifying a joint distribution $P(C,Y)$.  In our experiments we consider binary $C$ and $Y$, so that $P(C,Y)$ may be constructed as a 2 by 2 matrix, which involves 4 degrees of freedom.  We would instead like to reduce these to a single degree of freedom, which we term the bias parameter.

Intuitively, minimal bias occurs when the joint distribution factorizes, so that:
\begin{equation*}
    P_{min}(C,Y) = Pr(C)Pr(y)
\end{equation*}
The amount of bias may then be quantified by considering how much our specified $P$ deviates from this $P_{min}$, which we do using the entropy:

\begin{equation}
    H(P) = \sum_C \sum_Y P(C,Y) \log \left( \frac{P(C,Y)}{P_{min}(C,Y)}
    \right).
\end{equation}

\noindent We may thus define a $P_{max}$ via:

\begin{equation}
    P_{max} = \mathrm{argmax}_P \; H(P).
\end{equation}

\noindent This optimization is done with the three constraints:
\begin{align}
    \sum_C \sum_Y P_{max}(C,Y) &= 1 \\
    \sum_Y P_{max}(C,Y) &= Pr(C) \\
    \sum_C P_{max}(C,Y) &= Pr(Y)
\end{align}
Here, the marginals are estimated from the respective datasets by examining the relative frequencies of $C=1$ and $Y=1$.  In addition we have the positivity constraints:
\begin{equation}
    P_{max}(C,Y) \geq 0; \forall C,Y.
\end{equation}
We thus have an optimization problem with 4 degrees of freedom and 3 constraints, and may find $P_{max}$ using a one-dimensional line search.  We then obtain our biased $P$ by linearly interpolating between these two extremes:
\begin{equation}
    P_{bias} = (1 - bias) P_{min} + (bias) P_{max}
\end{equation}

\begin{figure*}[!t]
     \centering
     \begin{subfigure}[b]{0.32\linewidth}
     \centering
        \includegraphics[width=\textwidth]{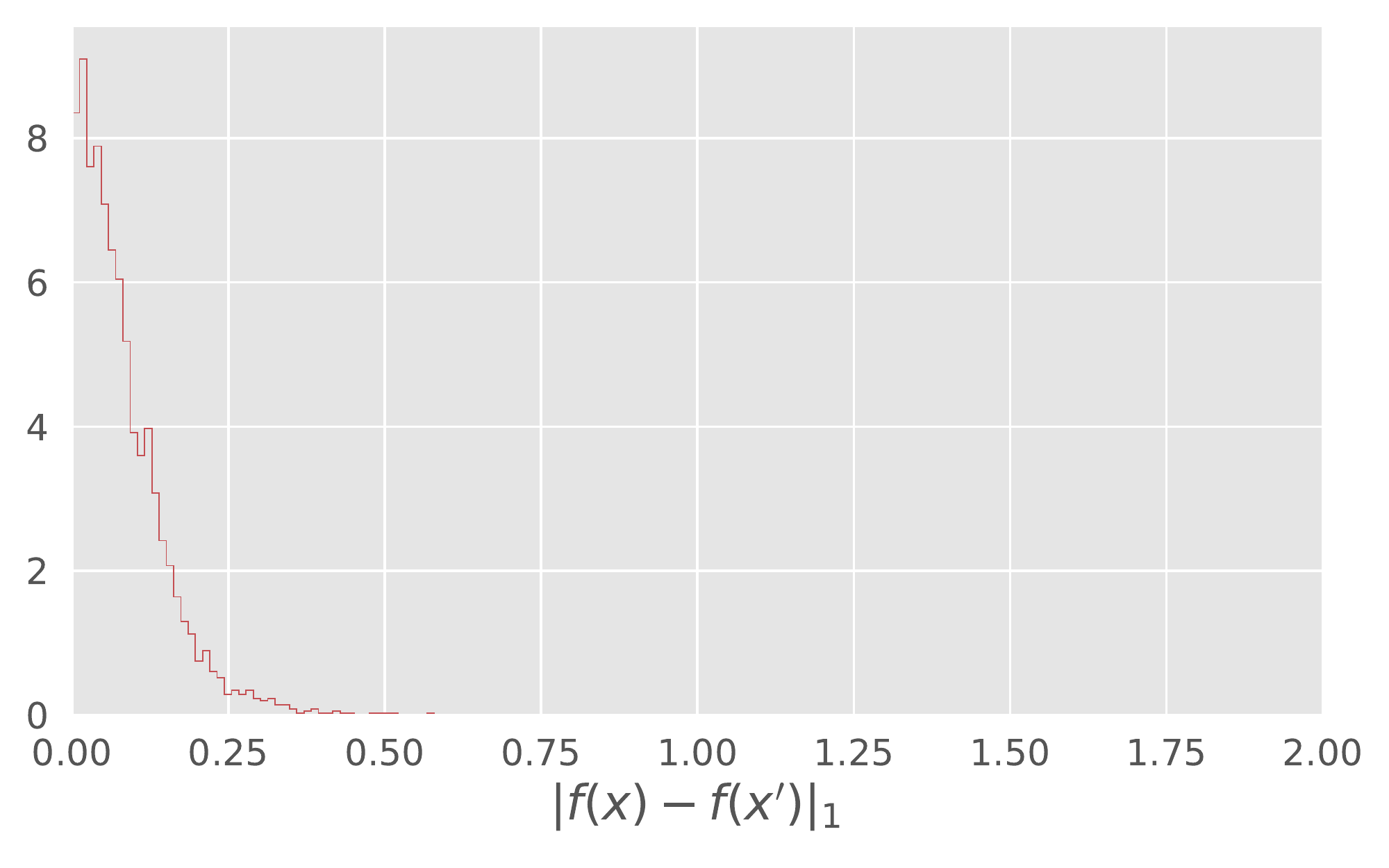}
        \caption{bias = 0}
        \label{fig:bias_0}
     \end{subfigure}
     \hspace{\fill}
     \begin{subfigure}[b]{0.32\linewidth}
     \centering
        \includegraphics[width=\textwidth]{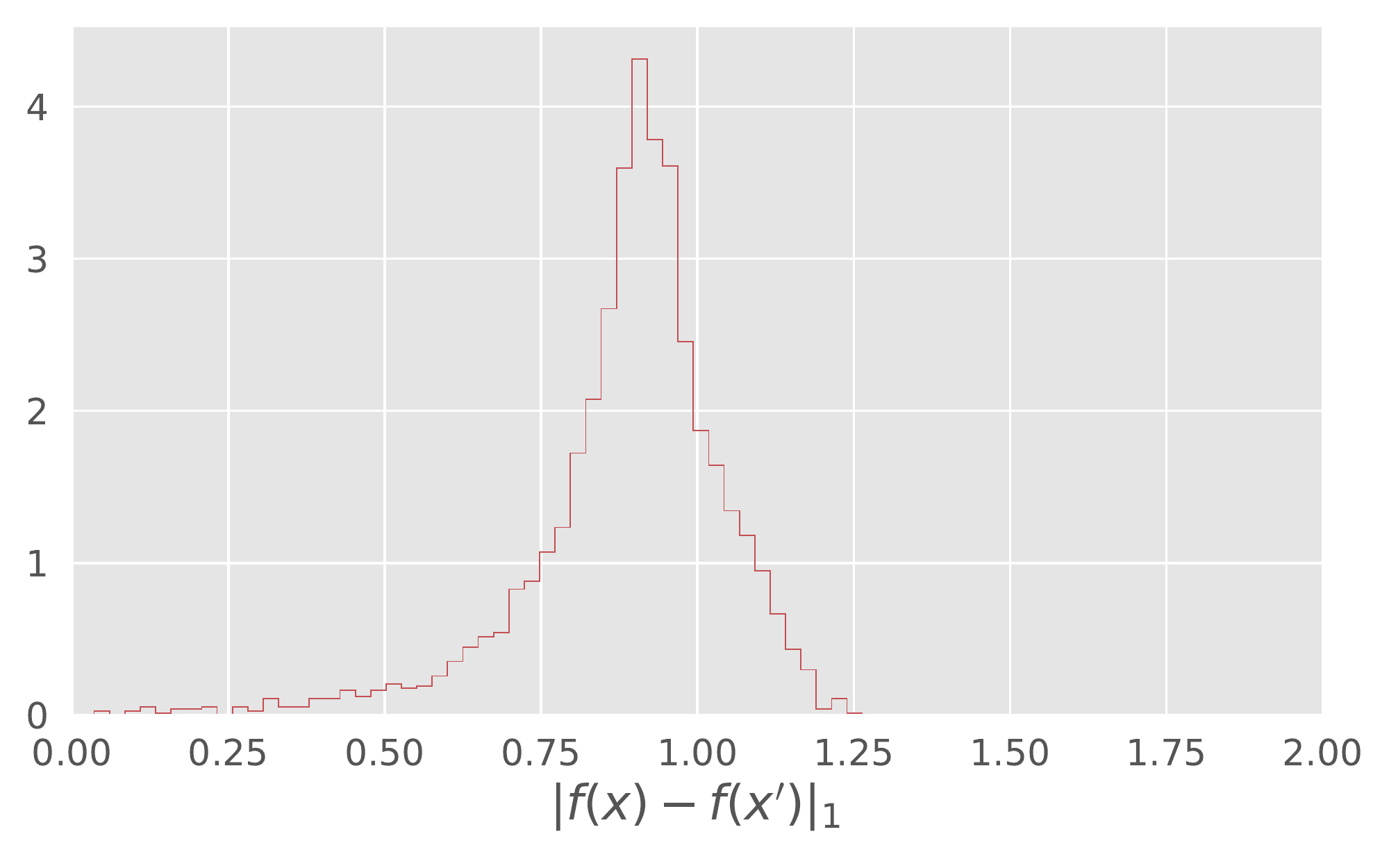}
        \caption{bias = 0.5}
        \label{fig:bias_0.5}
     \end{subfigure}
     \hspace{\fill}
    \begin{subfigure}[b]{0.32\linewidth}
    \centering
        \includegraphics[width=\textwidth]{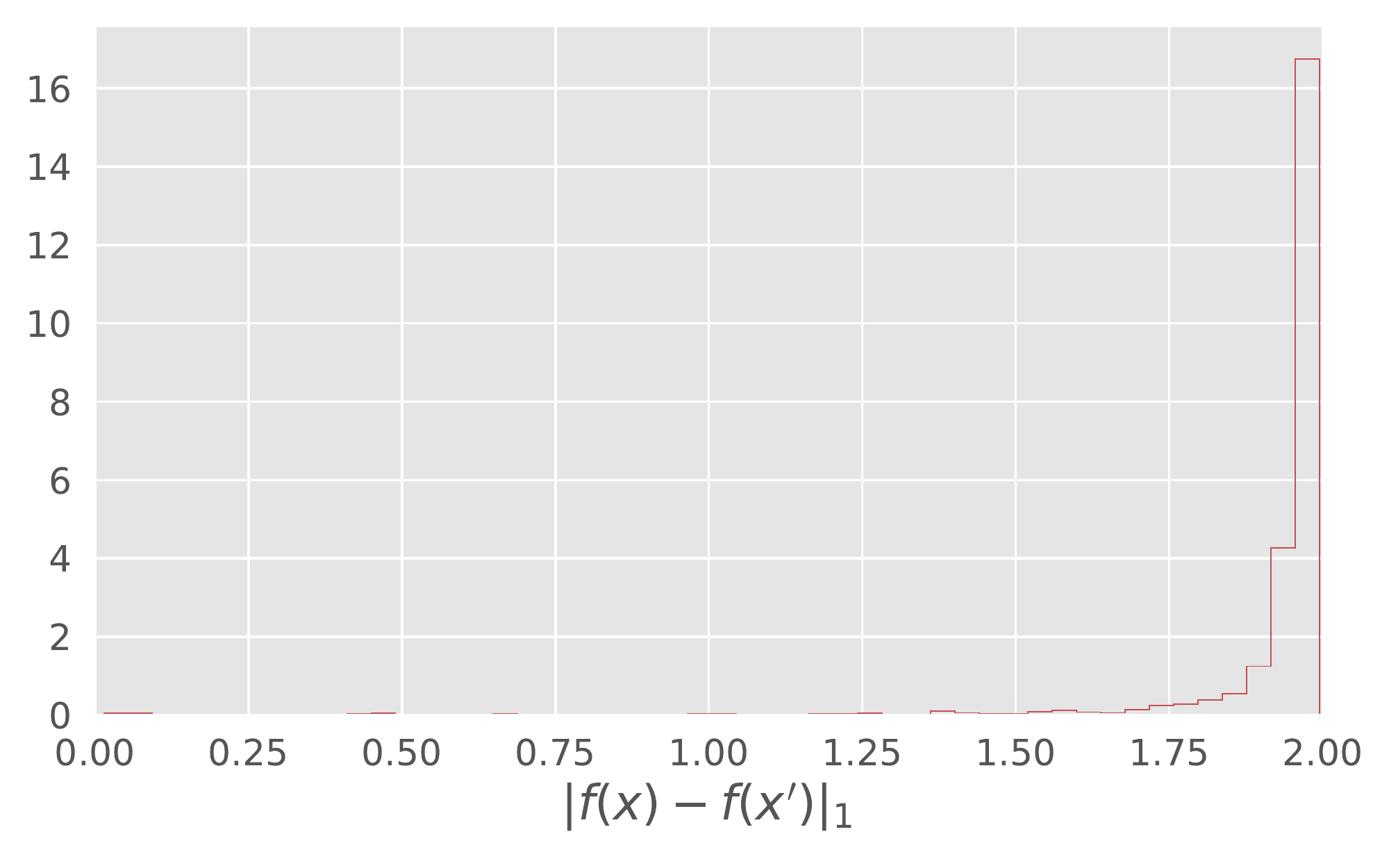}
        \caption{bias = 1}
        \label{fig:bias_1}
     \end{subfigure}
    \caption{\textbf{Controlling discrimination with tunable bias parameter.} Given access to a generative model, we can measure discrimination by evaluating the disparity in a model $f$'s outputs over a point $\textbf{x}$ and its counterfactual $\textbf{x}^{\prime}$.  The above histograms demonstrate that datasets with higher values of the bias parameter provoke higher rates of discrimination in models trained on them.}
    \label{fig:cff-bias}
    \vspace{-2mm}
    \end{figure*}

We can see the effect of adjusting bias in Figure \ref{fig:cff-bias}, demonstrating that our proposed mechanism does indeed provoke unfair treatment in models.

\subsubsection{Fusion Functions}\label{sec:supp:transforms}
In this section we consider how to fuse datasets together.  We consider two datasets $\hat{D}=\{\cdots (\hat{\mathbf{x}}_i,\hat{y}_i)\cdots\}$ and $\tilde{D}=\{\cdots (\tilde{\mathbf{x}}_j,\tilde{y}_j)\cdots\}$.  The inputs have dimensions $\hat{N}$ and $\tilde{N}$ respectively.

In fusing the datasets, we are slightly constrained: while we aim to mix the different patterns, they must also be distinct enough that the learning task can be performed. That is to say, all stochasticity and nonlinearity should come from the original datasets, whose patterns should be preserved (and thus, simultaneously learnable) under the fusion function. In our experiments, we thus consider two extremes of mixing:

\begin{itemize}[leftmargin=10pt, itemsep=0.1em, topsep=0pt]
    \item We concatenate the vectors $\hat{\mathbf{x}}_i$ and $\tilde{\mathbf{x}}_j$:
    \begin{equation}
        \textbf{x} = \textrm{concat}(\hat{\mathbf{x}}_i, \tilde{\mathbf{x}}_j)
    \end{equation}
    This corresponds to the setup in Kusner et al. [2017], in which the observable features $X$ may be partitioned into descendants and non-descendants of the protected variable $C$\footnote{That is to say, given a causal graph, certain features of $X$ may be children of the protected variable $C$}.  A simple linear projection is sufficient to separate these sources of variation:
    \begin{align}
    \hat{\mathbf{x}}_i = \hat{W} \textbf{x}&; \;\;\; \hat{W} = \mathrm{concat}(I_{\hat{N}}, 0)\\
    \tilde{\mathbf{x}}_j = \tilde{W} \textbf{x}&; \;\;\; \tilde{W} = \mathrm{concat}(0, I_{\tilde{N}})
    \end{align}
    \item As another extreme, we consider the case of perfect mixing by taking the outer product:
    \begin{equation}
        \textbf{x} = \mathrm{vec}(\hat{\mathbf{x}}_i \otimes \tilde{\mathbf{x}}_j).
    \end{equation}
    If $\hat{\mathbf{x}}_i$ and $\tilde{\mathbf{x}}_j$ each sum to 1, then we can completely isolate them from $\textbf{x}$ using a different linear projection:
    \begin{align}
    \hat{\mathbf{x}}_i = \hat{W} \textbf{x}&; \;\;\; \hat{W} = \frac{1}{\tilde{N}} (I_{\hat{N}} \otimes 1_{\tilde{N}})\\
    \tilde{\mathbf{x}}_j = \tilde{W} \textbf{x}&; \;\;\; \tilde{W} = \frac{1}{\hat{N}} (1_{\hat{N}} \otimes I_{\tilde{N}})
    \end{align}
    
    %TODO: add the linear projections here
    
\end{itemize}

\begin{table}[!t]
\centering
\caption{\textbf{Original datasets used in the synthetic pipeline}: information on the datasets from PMLB, along with the hyperparameters of the CVAEs used to model them.  A lower model score corresponds to a more "difficult" dataset to model. }
\resizebox{\linewidth}{!}{%
\begin{tabular}{c|c|c|c|c|c|c}
\toprule
dataset    & \# features & \# instances & \begin{tabular}[c]{@{}c@{}}\# gmm \\ components\end{tabular} & hidden size & depth & \begin{tabular}[c]{@{}c@{}}model\\ score\end{tabular} \\ \midrule[1.5pt]
backache   & 6           & 180          & 3                                                            & 186         & 6     & 0.545                                                 \\ \hline
credit     & 7           & 1000         & 7                                                            & 874         & 3     & 0.657                                                 \\ \hline
australian & 6           & 690          & 5                                                            & 977         & 5     & 0.721                                                 \\ \hline
wdbc       & 30          & 569          & 4                                                            & 367         & 1     & 0.815                                                 \\ \hline
magic      & 10          & 19020        & 2                                                            & 624         & 2     & 0.987                                                 \\ \bottomrule
\end{tabular}}
\label{tab:Datasets}
\end{table}

\subsubsection{Generative Models}

For ease of generating counterfactuals, we use Conditional Variational Autoencoders (CVAEs) to generate samples $X$ from a specified target $Y$.  We discard all categorical features, and retain only those which are real values.  Because the datasets used are tabular, we use a pre-processing strategy similar to \cite{CTGAN}, modelling each feature with a Gaussian Mixture Model (GMM).  This models the univariate feature distribution as a sum of distinct gaussian modes, with each mode assigning the feature a certain probability.  We may thus transform each feature into a vector of mode probabilities, and we concatenate these probability vectors to form our inputs $\textbf{x}$.  Since each probability vector sums to 1, this has the added advantage that the sum over $\textbf{x}$ will be a fixed integer, namely, the number of features, making it compatible with the outer product transform described in Section \ref{sec:supp:transforms}.

In addition to the depth and width of the encoder/decoder, we tune the number of components of the GMM when training our generative models.  To evaluate the final model, we sample a collection of $(x,y)$ pairs, and use them to train a simple classifer to predict $y$ from $x$.  We tune our parameters so as to maximize the accuracy of this classifier.  Intuitively, this results in a generative model that best preserves the original discriminative relationship inside a given dataset.  Model hyperparameters are shown in Table \ref{tab:Datasets}.

\subsubsection{Datasets Used}\label{sec:supp:dsets-used}

In determining our datasets, we aim to find a combination of stochastic and nonlinear patterns.  To quantify the complexity of these patterns, we measure the maximum accuracy reached by a simple classifier after tuning.  Intuitively, if a model is unable to obtain a high score, that dataset constitutes a more difficult challenge.  We perform this analysis on all of the binary classification datasets in PMLB~\cite{Olson2017PMLB}, and select those listed in Table \ref{tab:Datasets}, since they represent a spread of complexity.  Note that these datasets are distributed under an MIT License.

Note that, in selecting and training models on these datasets, we retain only those features which are continuously valued, in order to match our generative model setup.

\subsubsection{Specific Settings used in Section \ref{sec:insights}}\label{sec:supp:specific-settings-for-insights}

To generate the plots in Section \ref{sec:insights} of the main paper, we generate synthetic datasets according to the following parameter grid:

\begin{itemize}[leftmargin=10pt, itemsep=0.1em, topsep=0pt]
    \item We select the \textit{bias} parameter (Section \ref{sec:supp:bias-param}) from the following choices: $[0,0.25,0.5,0.75,1.0]$.
    \item We select the fusion function (Section \ref{sec:supp:transforms} from the set $[\textrm{concat}, \textrm{outer}]$.
    \item We select all pairs of datasets from Table \ref{tab:Datasets}.
\end{itemize}

\noindent We train fully connected neural networks for our auxiliary and target models, using randomly sampled widths (8-128 units) and depths (1-4 layers).

\subsubsection{Constructing Ground-Truth Labels and Models}
The main utility of using generative models is that we may simultaneously sample both an instance $\textbf{x}$ and its counterfactual $\textbf{x}^{\prime}$.  Accordingly, by passing $\textbf{x}$ and $\textbf{x}^{\prime}$ through the target model $f_{tar}$, we may use the $l_1$-distance between the model outputs as a measure of discrimination, which we refer to as the Individual Fairness Score (IFS):
\begin{equation}
    \textrm{IFS} = |f_{tar}(\textbf{x}_i) - f_{tar}(\textbf{x}_j)|_1
\end{equation}

Figure \ref{fig:cff-bias} demonstrates how this CFF score varies as the parameters of our pipeline are adjusted.

To obtain binary labels, we define a threshold on the CFF score, which we compute using the variance of the CFF score in the absence of any bias.  Above this threshold, we say that the model is discriminating against the individual $\textbf{x}$ on the basis of $c$.

Furthermore, knowing the form of the fusion functions as described in Section \ref{sec:supp:transforms}, we may constrain the input layers of an MLP to separate the different patterns contained inside the synthetic data.  Models which project out the $C$ dependence become perfectly fair target models.  Models which project out the $Y$ dependence become perfectly disentangled auxiliary models.  %\Christopher{not sure what you mean by "project out" here..}\Giuseppe{Refer to section B.2, where we talk about different projection matrices that disentangle the sources of variaition.}

\subsection{Sources for the Experiments on Real Datasets}

The Adult and Bank datasets were obtained from IBM Fairness 360 \cite{AIFairness360}, and the SSL dataset was obtained from the Chicago Data Portal, which may be accessed via \href{https://data.cityofchicago.org/Public-Safety/Strategic-Subject-List-Historical/4aki-r3np}{this hyperlink.}

\subsection{Full Experiment Results for Real Datasets}
\begin{figure*}[t]
     \centering
     \begin{subfigure}{\linewidth}
     \centering
     \begin{subfigure}{0.13\linewidth}
        \includegraphics[width=\textwidth]{figs/adult_fta.pdf}
     \end{subfigure}
     \begin{subfigure}{0.13\linewidth}
        \includegraphics[width=\textwidth]{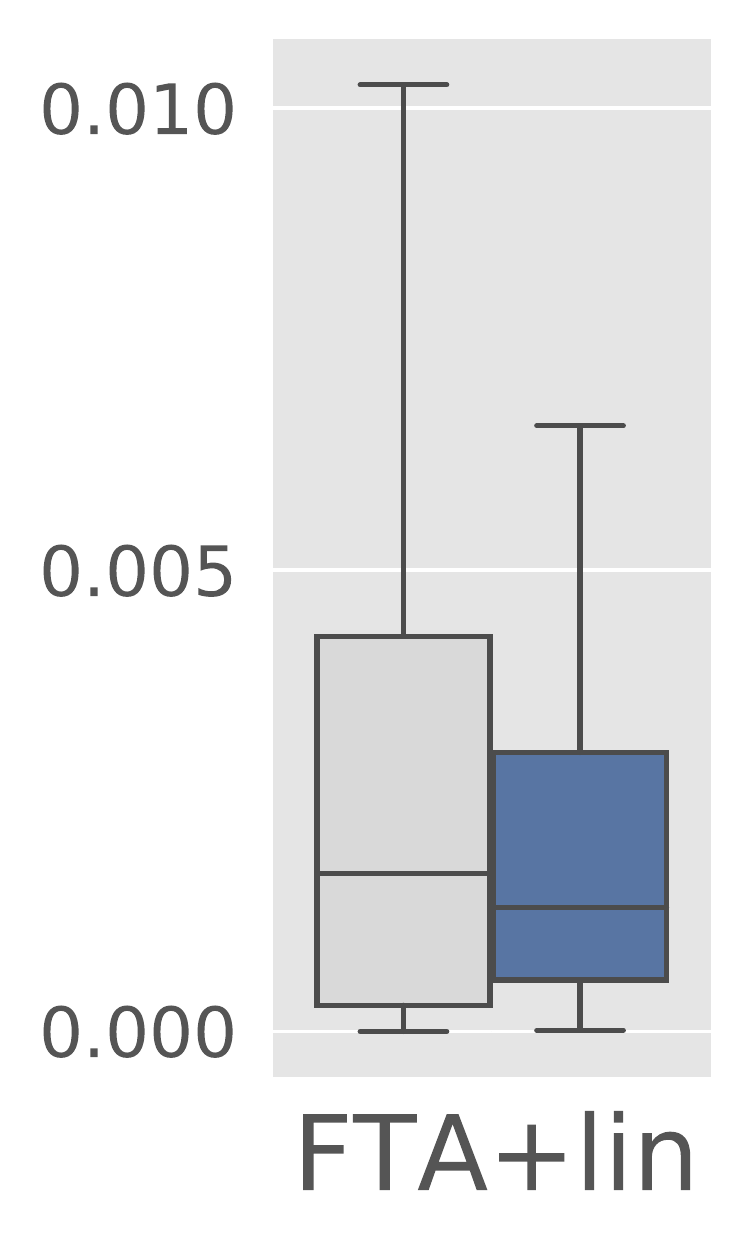}
     \end{subfigure}
     \begin{subfigure}{0.13\linewidth}
        \includegraphics[width=\textwidth]{figs/adult_unfair_map.pdf}
     \end{subfigure}
     \begin{subfigure}{0.13\linewidth}
        \includegraphics[width=\textwidth]{figs/adult_fliptest.pdf}
     \end{subfigure}
     \begin{subfigure}{0.13\linewidth}
        \includegraphics[width=\textwidth]{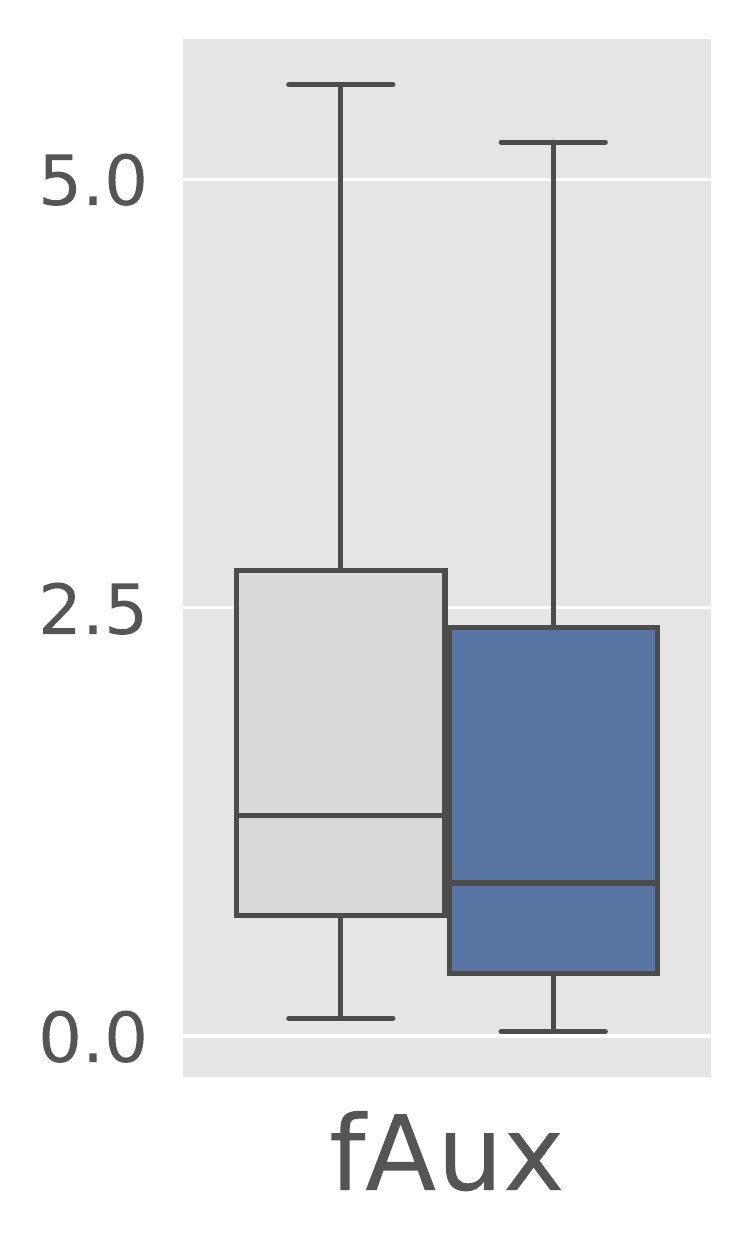}
     \end{subfigure}
     \begin{subfigure}{0.13\linewidth}
        \includegraphics[width=\textwidth]{figs/adult_norm_faux.pdf}
     \end{subfigure}
     \begin{subfigure}{0.13\linewidth}
        \includegraphics[width=\textwidth]{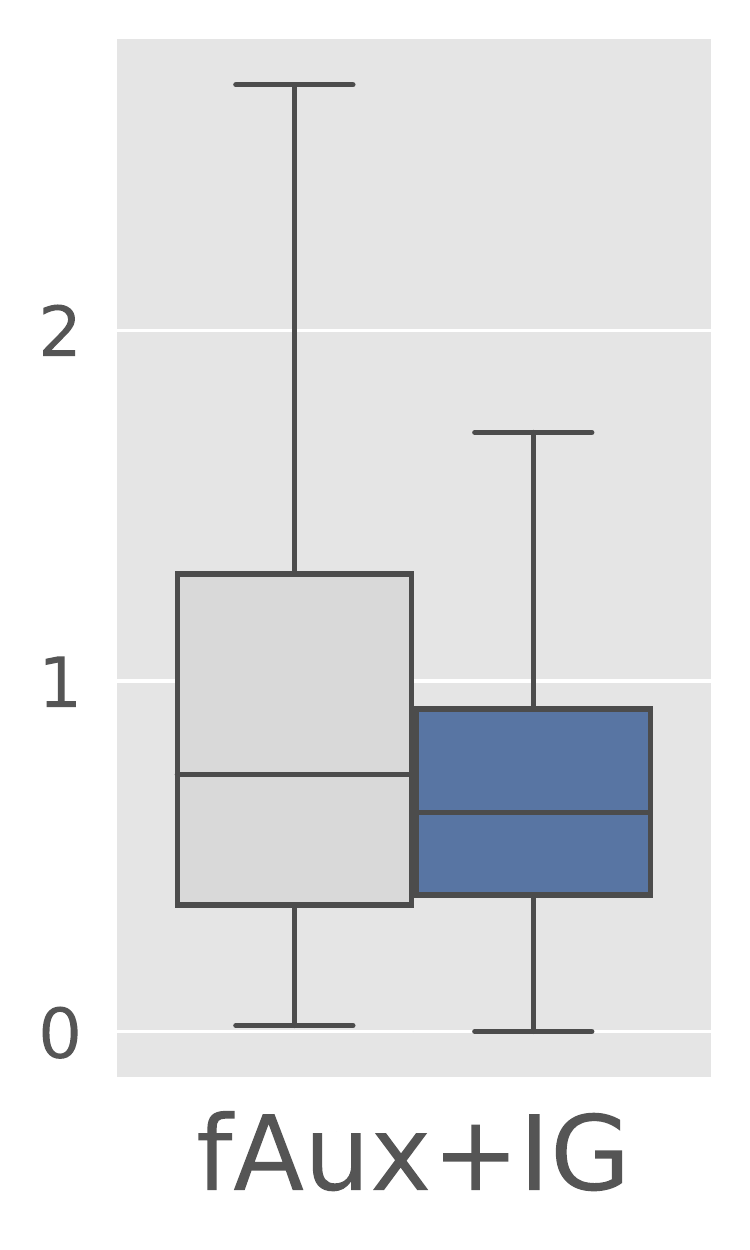}
     \end{subfigure}
     \caption{Adult Dataset}
     \end{subfigure}
     
     \begin{subfigure}{\linewidth}
     \centering
     \begin{subfigure}{0.13\linewidth}
        \includegraphics[width=\textwidth]{figs/bank_fta.pdf}
     \end{subfigure}
     \begin{subfigure}{0.13\linewidth}
        \includegraphics[width=\textwidth]{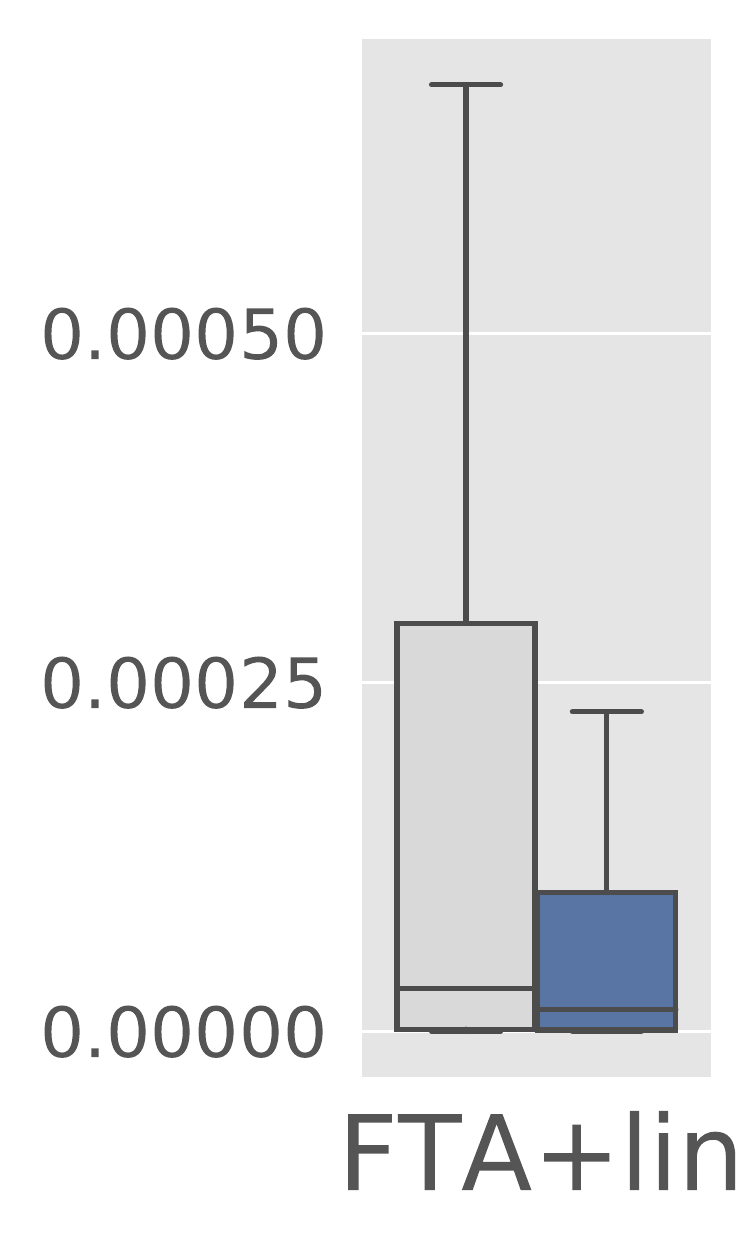}
     \end{subfigure}
     \begin{subfigure}{0.13\linewidth}
        \includegraphics[width=\textwidth]{figs/bank_unfair_map.pdf}
     \end{subfigure}
     \begin{subfigure}{0.13\linewidth}
        \includegraphics[width=\textwidth]{figs/bank_fliptest.pdf}
     \end{subfigure}
     \begin{subfigure}{0.13\linewidth}
        \includegraphics[width=\textwidth]{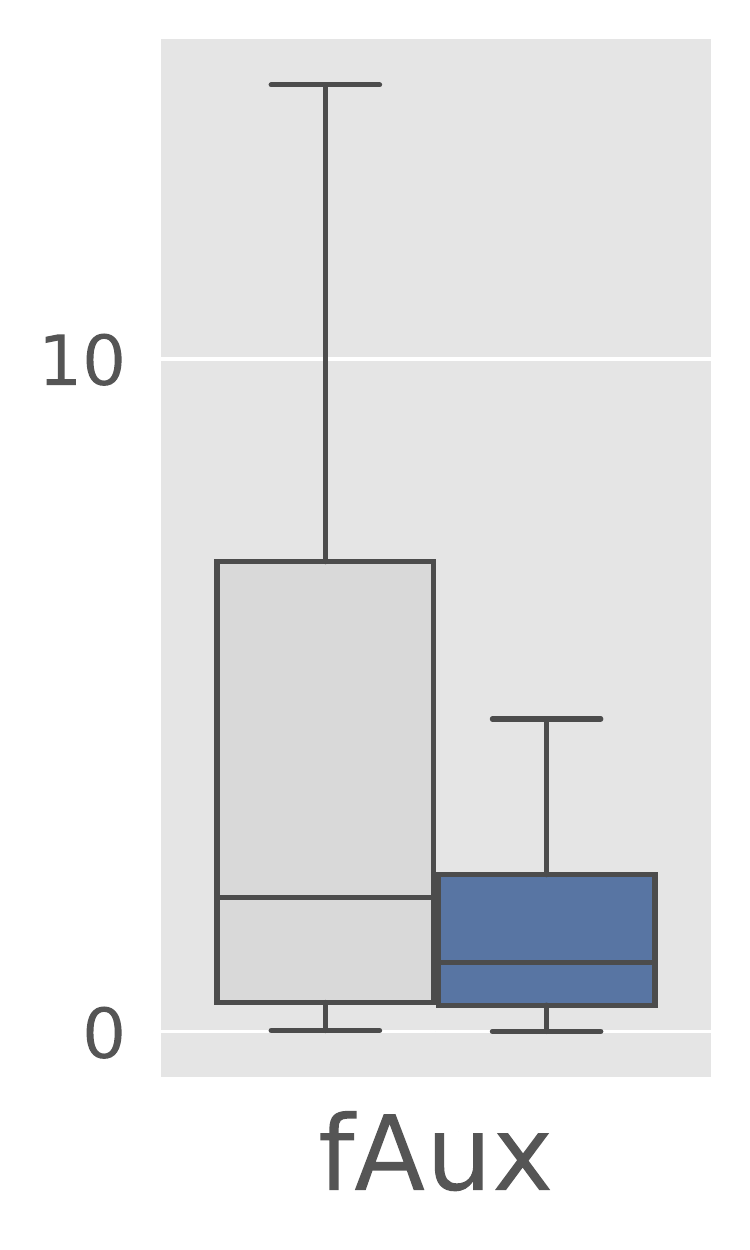}
     \end{subfigure}
     \begin{subfigure}{0.13\linewidth}
        \includegraphics[width=\textwidth]{figs/bank_norm_faux.pdf}
     \end{subfigure}
     \begin{subfigure}{0.13\linewidth}
        \includegraphics[width=\textwidth]{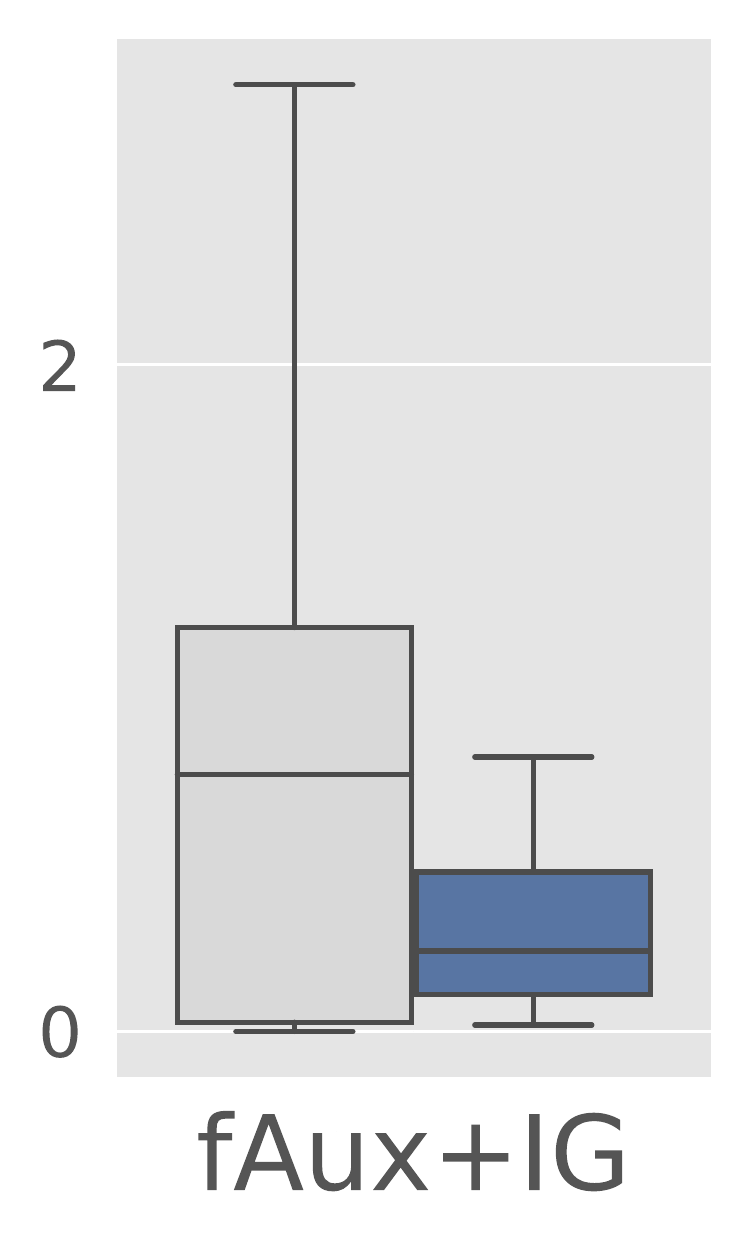}
     \end{subfigure}
     \caption{Bank Dataset}
     \end{subfigure}
     
     \begin{subfigure}{\linewidth}
     \centering
     \begin{subfigure}{0.13\linewidth}
        \includegraphics[width=\textwidth]{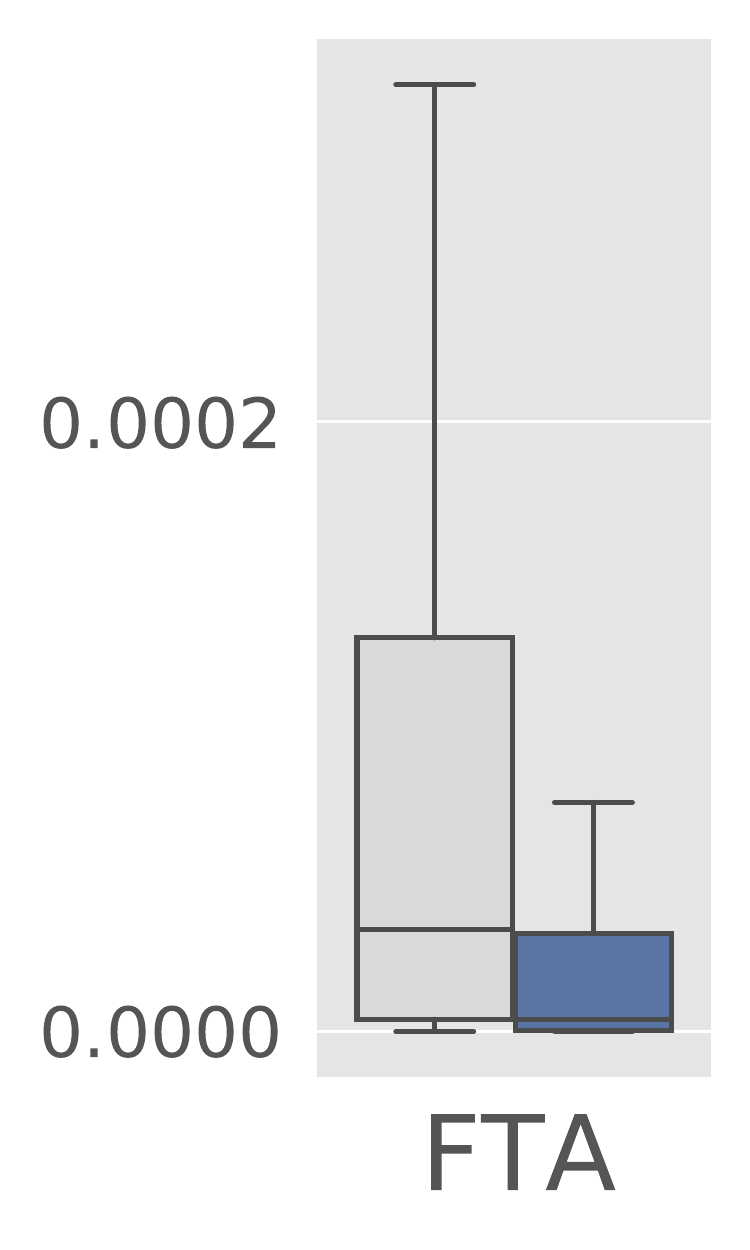}
     \end{subfigure}
     \begin{subfigure}{0.13\linewidth}
        \includegraphics[width=\textwidth]{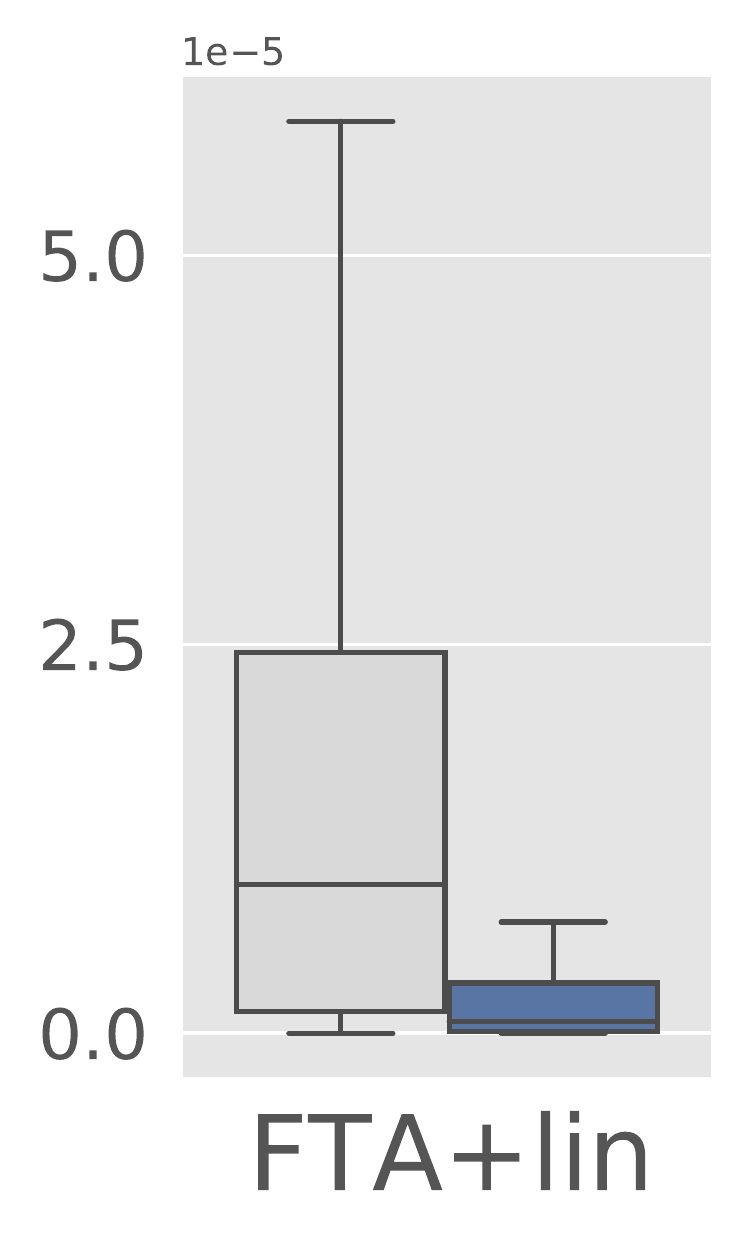}
     \end{subfigure}
     \begin{subfigure}{0.13\linewidth}
        \includegraphics[width=\textwidth]{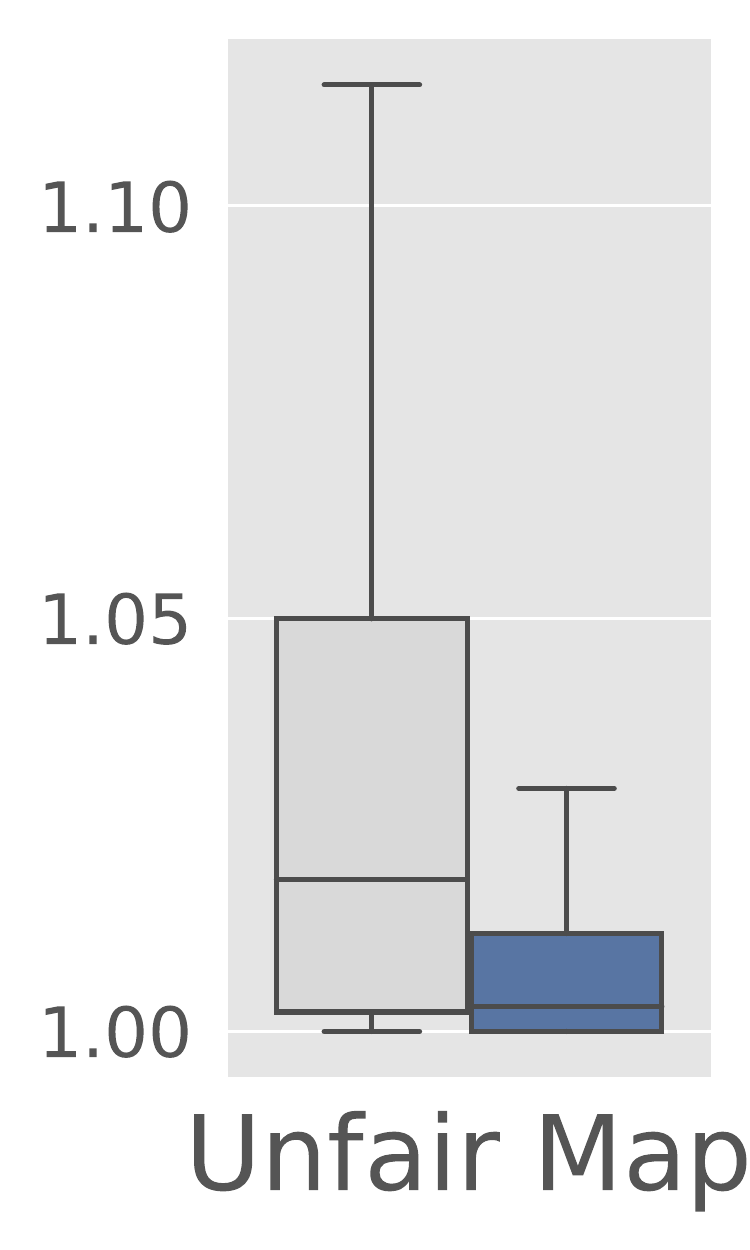}
     \end{subfigure}
     \begin{subfigure}{0.13\linewidth}
        \includegraphics[width=\textwidth]{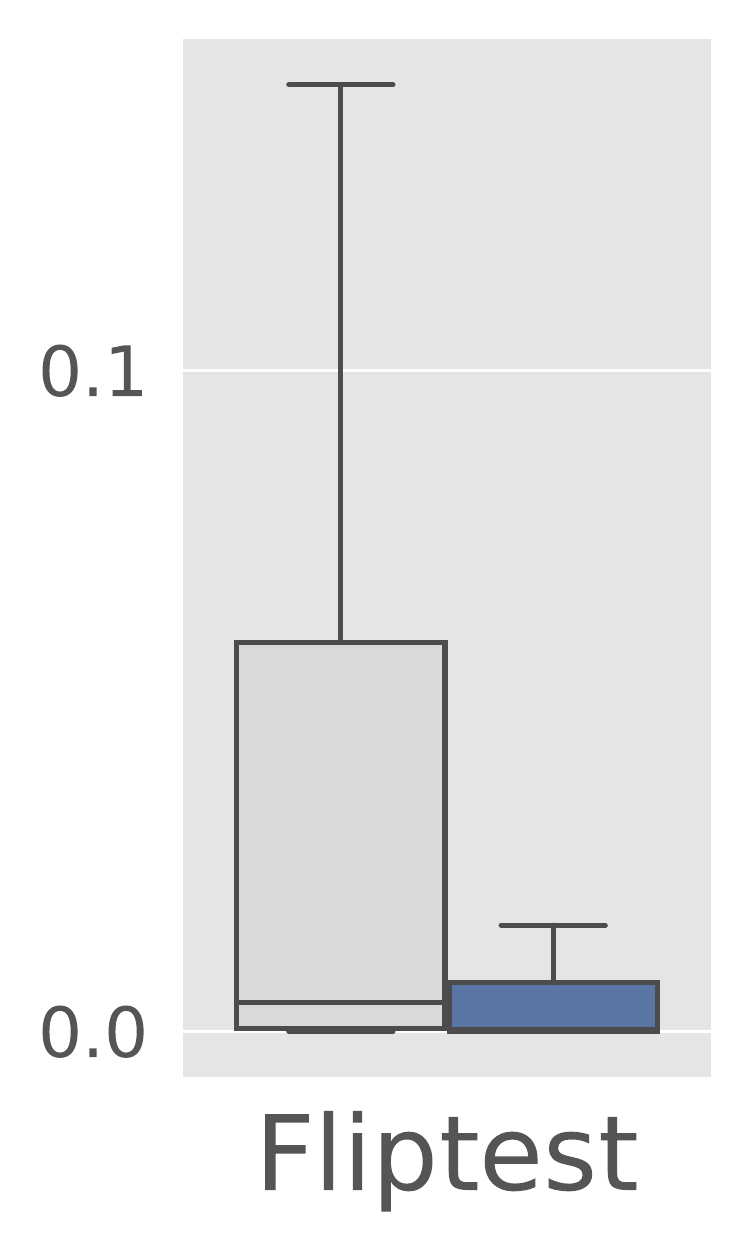}
     \end{subfigure}
     \begin{subfigure}{0.13\linewidth}
        \includegraphics[width=\textwidth]{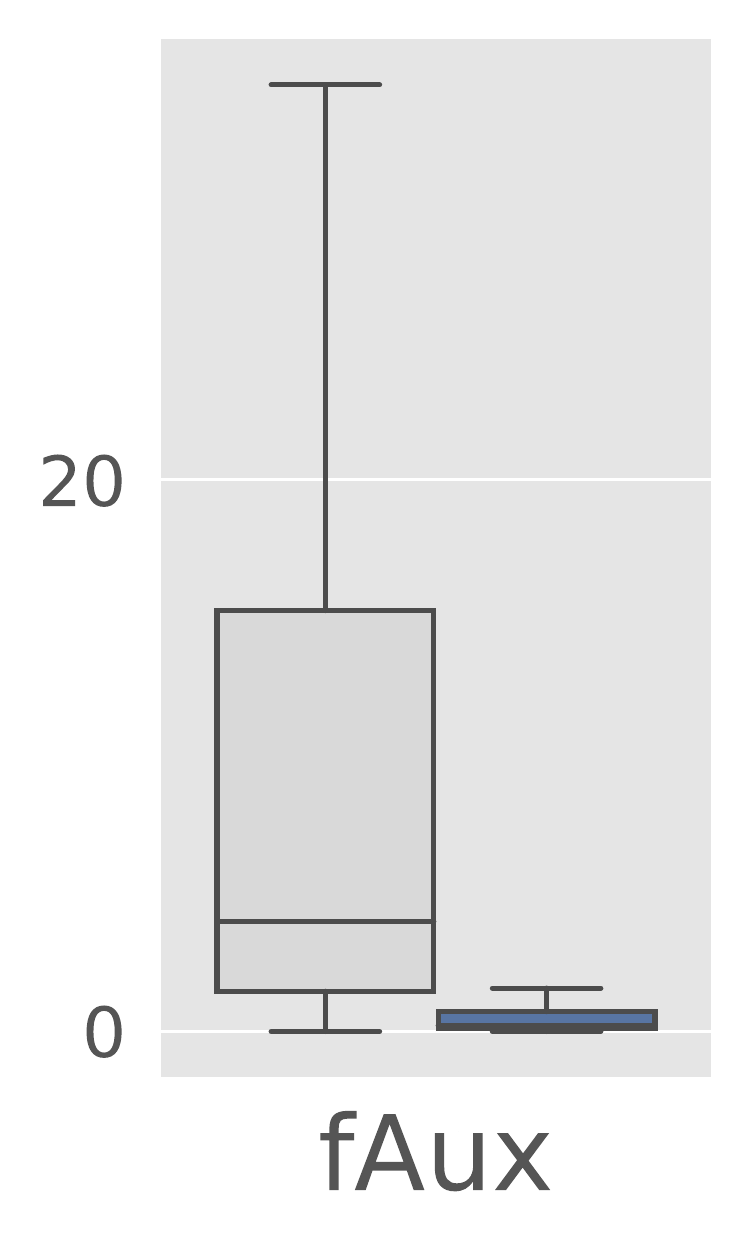}
     \end{subfigure}
     \begin{subfigure}{0.13\linewidth}
        \includegraphics[width=\textwidth]{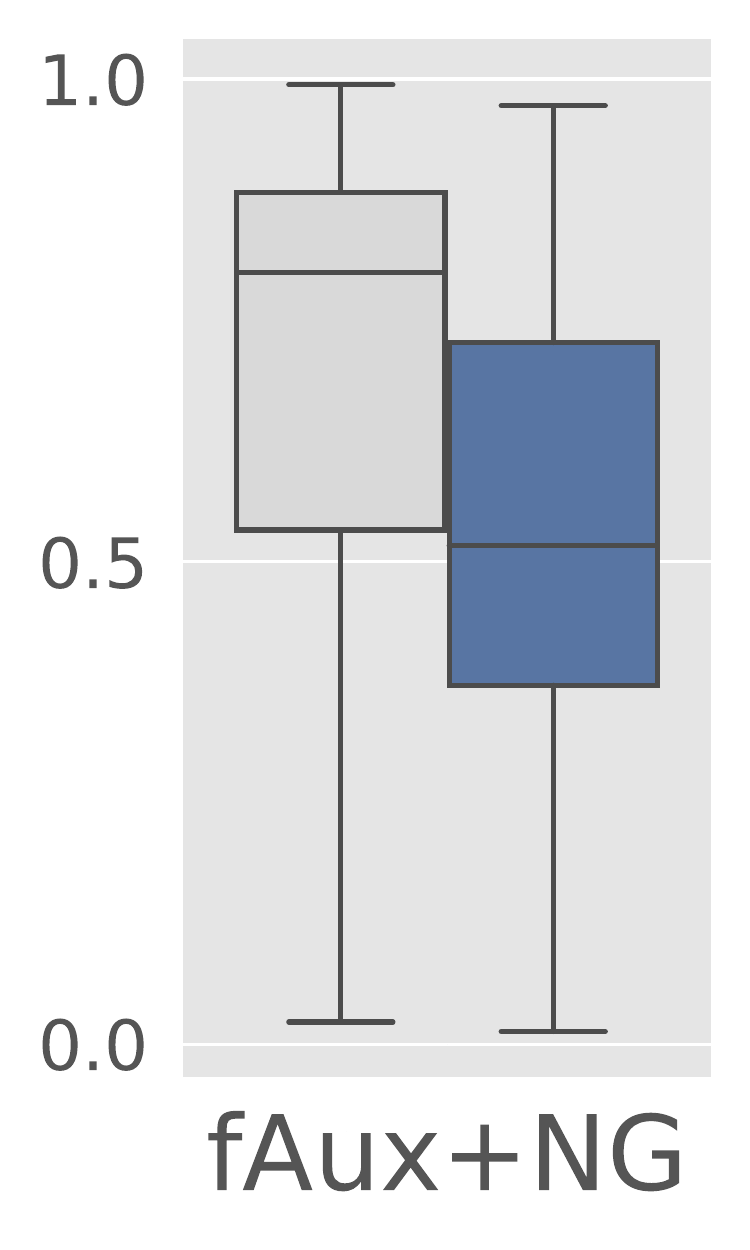}
     \end{subfigure}
     \begin{subfigure}{0.13\linewidth}
        \includegraphics[width=\textwidth]{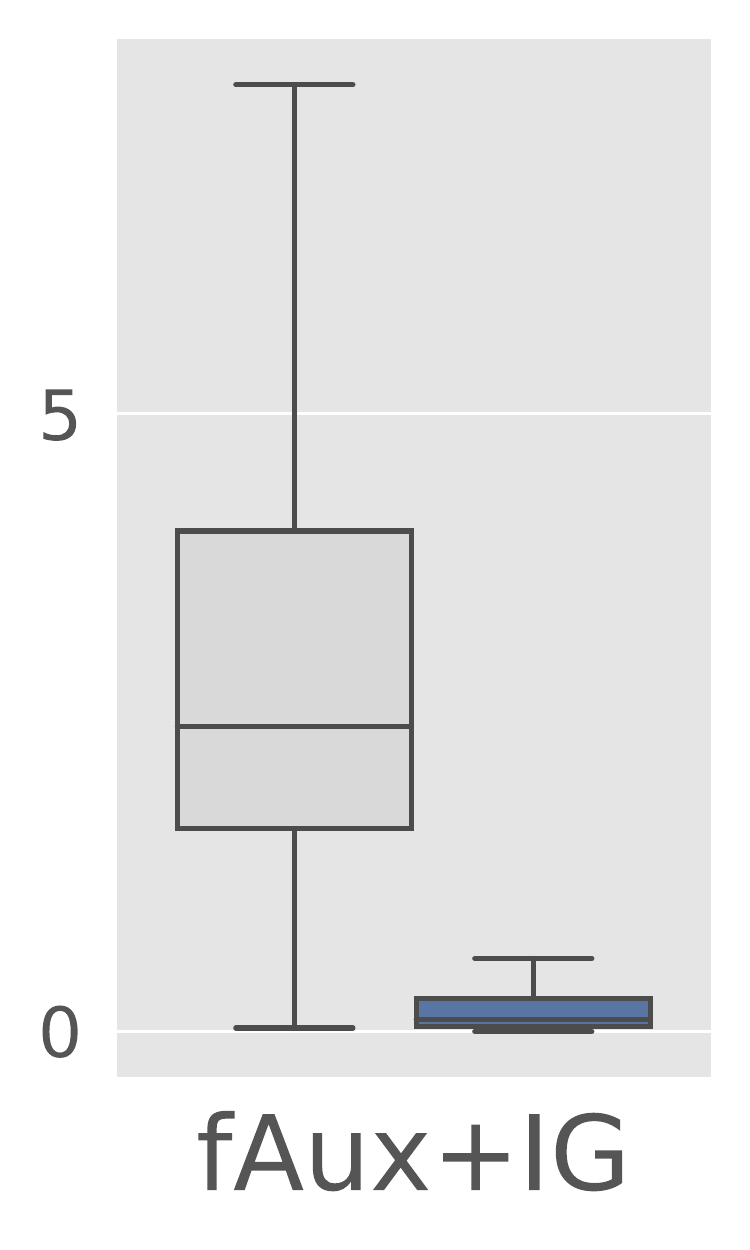}
     \end{subfigure}
     \caption{Chicago SSL Dataset}
     \end{subfigure}
         \caption{\textbf{Performance Comparison among Individual Fairness Testing Methods on Real Datasets.} For each testing algorithm, we show the predicted unfairness score for both unfair model~(grey box) and fair model~(blue box). Greater difference (between the boxes) shows better performance. }
    \label{fig:real_data_exp_full}
\end{figure*}

Due to the limited space in the main paper, we were unable to show the complete experiment results. Here, we show the full version of the performance comparison on real datasets. Figure~\ref{fig:real_data_exp_full} shows the complete plots on the three datasets.

\begin{table}[t]
\caption{\textbf{Reliability of Transparency Reports} Here, we compare the candidate testing methods' ability for flagging surrogate features of protected attributes through ranking metric NDCG.}
\resizebox{\linewidth}{!}{%
\begin{tabular}{l|c|c|c}
\toprule
                      & Adult        & Chicago SSL & Bank  \\ \midrule
Protected attribute    & Sex (Binary) & Race (Binary) & Age (Binarized) \\ \midrule[1.5pt]
FlipTest               & 
0.810 $\pm$ 0.150            & 
0.840 $\pm$ 0.136            &
0.752 $\pm$ 0.108
\\ \midrule[1.5pt]
fAux                    & 
0.802 $\pm$ 0.058            & 
0.796 $\pm$ 0.098           &
0.798 $\pm$ 0.047
%\\ \midrule
%fAux+SG                 & 
%0.714 $\pm$ 0.093            & 
%0.829 $\pm$ 0.127             &
%0.726 $\pm$ 0.092
\\ \midrule
fAux+NG                 & 
0.807 $\pm$ 0.105            & 
0.884 $\pm$ 0.119             &
0.789 $\pm$ 0.058
\\ \midrule
fAux+IG                 & 
\textbf{0.896 $\pm$ 0.037}            & 
\textbf{0.958 $\pm$ 0.006}    &
\textbf{0.814 $\pm$ 0.058}
\\ \bottomrule
\end{tabular}}
\label{table:real-data-res}
\end{table}
\begin{table}[t]
\centering
\caption{\textbf{Resource Consumption Comparison between FlipTest and fAux.} The fAux variants share the same basic architecture and so we use fAux to represent all variants.}
\resizebox{\linewidth}{!}{%
\begin{tabular}{cc|c|c|cc}
\toprule
&&Resource&FlipTest&fAux&\\
\midrule[1.5pt]
&\multirow{2}{*}{Adult}& \# Parameters  & 220 $\pm$ 187 K  & 13 $\pm$ 8 K& \\
&& Training time & 1012 $\pm$ 532 s & 12 $\pm$ 5 s &\\
\midrule
&\multirow{2}{*}{Chicago SSL}& \# Parameters & 154 $\pm$ 150 K & 7 $\pm$ 6 K &\\
&& Training time & 1086 $\pm$ 580 s & 90 $\pm$ 23 s &\\
\midrule
&\multirow{2}{*}{Bank}& \# Parameters & 206 $\pm$ 159 K  & 13 $\pm$ 10 K &\\
&& Training time & 444 $\pm$ 204 s & 15 $\pm$ 4 s &\\
\bottomrule         
\end{tabular}}
\label{table:comp-eff}
\end{table}

%\Simon{Still need to know what the numbers in table 2 actually are.  Should be added to legend.}
% table 2 only has NDCG score
%What is that?  Did not see this mentioned in text.  THere is a discussion  under 4.1.4 but nothing concrete.
% ye, missed it. I will add NDCG description in 4.1.4. added

\subsection{Explaining Discrimination With Transparency Reports }
\label{sec:comparisons_real-dsets}

As the real datasets do not have corresponding generative models that produce the data, we cannot compute IFS as the ground-truth label and so direct performance comparison on individual fairness testing is infeasible.  In the main body of the paper, we opted to compare the scores output by the fairness test on fair and unfair models.  However, another way to evaluate a fairness test is to examine the explanations it provides for flagging discrimination.  In this section, we examine how reliably fAux can generate such explanations.

In 
\cite{black2019fliptest} the authors construct transparency reports that rank features based on their contribution to an unfair decision.  To generate scores for the different tests, we collect all $N_0$ instances in the dataset for which $c=0$, to define a set $S_0$.  For Fliptest, we use the generator $G$ to compute the following vector:
\begin{equation}
\frac{1}{N_0}\sum_{\textbf{x} \in S_0} (\textbf{x} - G(\textbf{x}))
\end{equation}
The Fliptest feature scores are then given by the absolute value of this vector.  While fAux does not use an explicit Generator, it does define a transformation of features through the auxiliary model $f_{aux}$ \eqref{eq:penrose-invers}.  We can thus compute transparency reports from the following vector:

\begin{equation}
\frac{1}{N_0}\sum_{\textbf{x} \in S_0} \nabla f_{aux}(\textbf{x})
\end{equation}

\noindent Once again, the feature score is given by the absolute values of this vector.

To quantitatively compare these scores, we use a ranking metric, the Normalized Discounted Cumulative Gain~(NDCG).  Our ground truth ranking is based on an estimate of the mutual information between each feature and the protected variable $C$.  We use a nonparametric entropy-estimator based on \cite{Ross2014MutualIB}, which is available through scikit-learn.  Table~\ref{table:real-data-res} then compares the NDCG scores for the different test rankings.  Note that, for categorical features that are one-hot encoded, we compute an aggregate score for the feature by taking the mean of the scores of the one-hot components.

%Removed the citation to scikit-learn
%\cite{scikit-learn}

We note fAux+IG shows remarkably better performance than FlipTest in terms of explaining the reason for unfair treatment.  In addition, we may compare the computational efficiency of both approaches.  Table~\ref{table:comp-eff} shows the computation resources used for the previous experiment on the real datasets. We note that the proposed fAux framework is remarkably efficient. It uses  $\sim$5\% of the parameters and 10\% of the training time of FlipTest on both datasets. In combination with the results from Table~\ref{table:synth-data-res} and \ref{table:real-data-res}, this highlights the significant advantages of fAux in terms of both effectiveness and efficiency. %We attribute this to the success of \cite{sundararajan2017axiomatic}

\begin{figure*}[t]
     \centering
     \begin{subfigure}[b]{0.49\linewidth}
     \centering
        \includegraphics[width=0.95\textwidth]{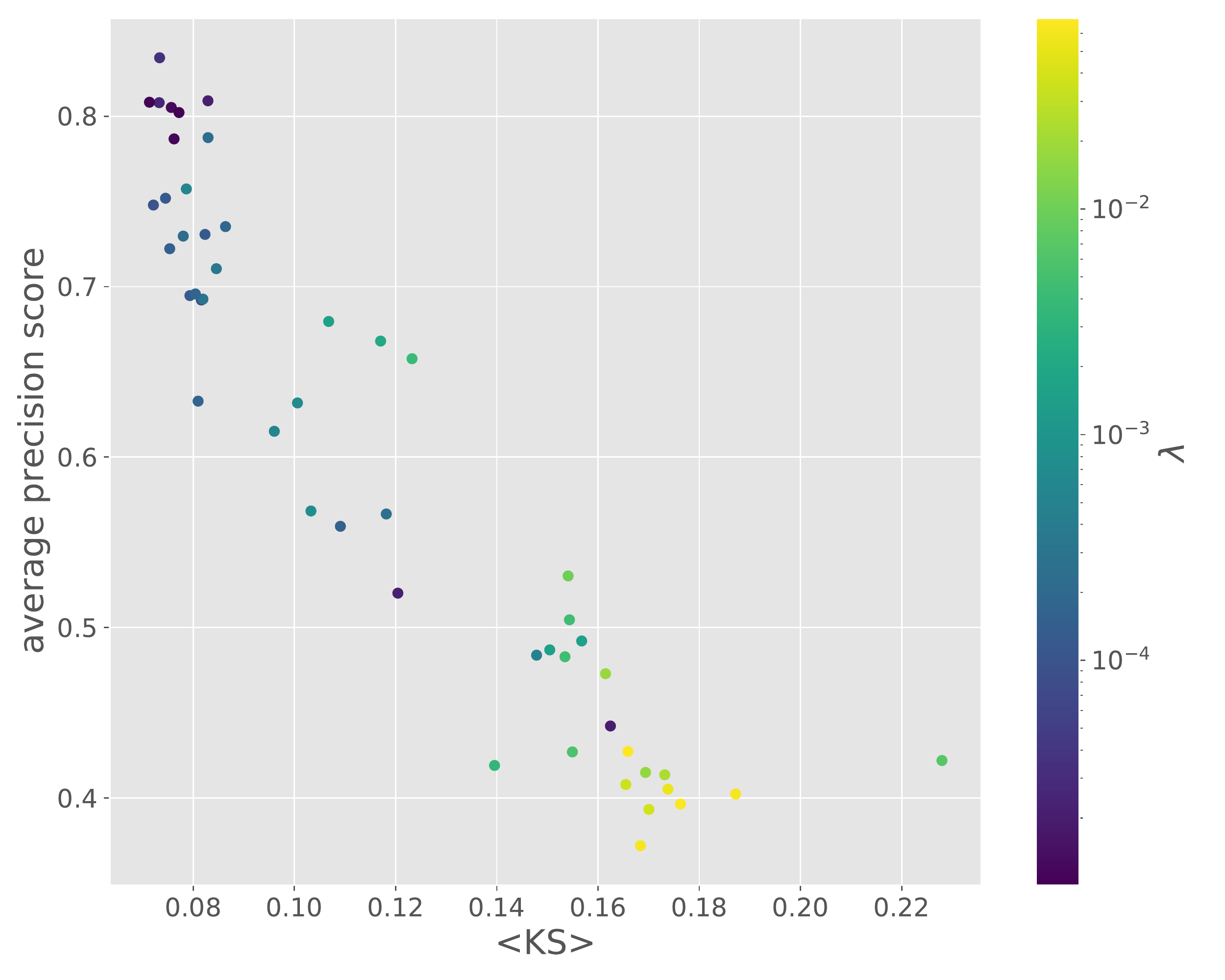}
        \caption{Average KS statistic}
        \label{fig:gan-metric-ks}
     \end{subfigure}
     \hspace{\fill}
    \begin{subfigure}[b]{0.49\linewidth}
    \centering
        \includegraphics[width=0.95\textwidth]{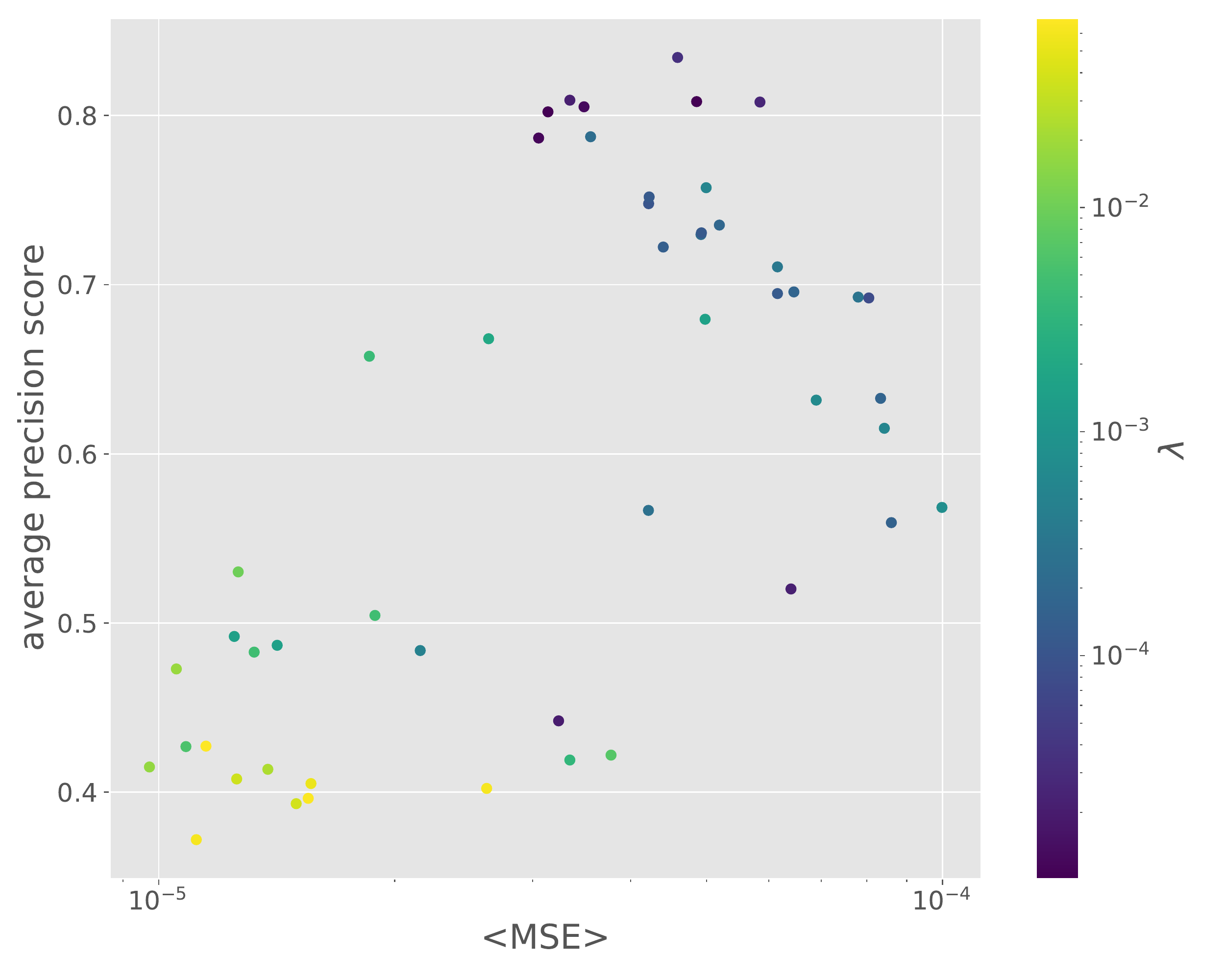}
        \caption{Average MSE}
        \label{fig:gan-metric-mse}
     \end{subfigure}
    \caption{\textbf{Correlation between GAN evaluation metrics and fairness test performance.} Given the high sensitivity of GANs to their hyperparameters, it is important to find evaluation criteria to tune them.  The authors of Fliptest suggest two, average KS and average MSE, and we compare how these metrics are related to the average precision of the fliptest.  The plots show the effect of using different values of the transport weight $\lambda$.  We find that, while average KS does correlate well with average precision, average MSE does not.  }
    \label{fig:gan-metrics}
    \vspace{-3mm}
    \end{figure*}

\subsection{Hyperparameters for the Fairness Tests}\label{sec:hyperparams-for-tests}

In this section, we review the hyperparameters used for the different fairness tests for both the synthetic and real datasets.   All experiments were distributed on our internal cluster which contain NVIDIA DGX-1 with 8X NVIDIA Tesla® V100 32 GB/GPU.  The cluster runs Red Hat Enterprise Linux Server release 7.9, and our experiments employ pytorch version 1.9.0.  Unless otherwise stated, all stochastic algorithms (ex minibatch gradient descent, parameter initialization, etc) use a random seed of 0.  

\subsubsection{Hyperparameters for Constructing Fair Models}

For the real datasets, without knowledge of the fusion-function, we employ \cite{OneNetworkFairness} to train fair models using adversarial regularization.  There were two parameters to tune:

\begin{itemize}
    \item The regularization strength $\alpha$.  This is the initial relative weight given to the regularizer over the loss function, and this weight exponentially decays between epochs.
    
    \item Adversarial training can sometimes be unstable, and so for every iteration that we trained the target model, we train the adversarial model for $n_{adv}$ epochs.  
\end{itemize}

\noindent \cite{OneNetworkFairness} was quick to use on our datasets, and so we tuned these values manually.  In particular, we tuned the regularization strength on our synthetic datasets since they offered a more accurate assesment of the tradeoff between accuracy and fairness.  We found that an $\alpha$ of 100 yielded good results.  We tuned $n_{adv}$ in a similar fashion, and obtained a value of 3.  We used these values when training models on all of the real datasets.

\subsubsection{Hyperparameters for fAux}

We use MLPs for all auxiliary models, varying the depth and width across different experimental runs.  We use a fixed batch size of 64 across all experiments, using the ADAM optimizer and a learning rate of 0.001 with early stopping.

%MODIFIED: removed ADAM citations
%\cite{kingma2017adam}

For the real datasets, we additionally employ \cite{OneNetworkFairness}, so that the resulting auxiliary model attends to those features that are most strongly correlated with the protected attribute.  We employ the same settings that we use when constructing the fair models.

\subsubsection{Hyperparameters for Unfair Map}

The Unfair Map \cite{maity2021statistical} uses a gradient-flow attack restricted to a sensitive subspace in order to test individual fairness.  Tensorflow code for this paper was publically released through OpenReview\footnote{Link \href{https://openreview.net/forum?id=z9k8BWL-_2u}{here}.}, and our own (pytorch) implementation is based off this.  There are three hyperparameters to consider:

\begin{itemize}
    \item The regularization strength, which is responsible for restricting the attacks to the sensitive subspace.
    
    \item The learning rate, which is used to make the gradient updates in the attack.  
    
    \item The number of steps used in the attack.
\end{itemize}

\noindent In our experiments, we use the same values for these parameters that are provided in the reference implementation.  In addition to these parameters, this test requires a fair metric, which is constructed from a logistic regression model.  To determine the sensitivity of this approach to the fair metric used, we train the logistic regression model using scikit-learn using different random seeds.  For the synthetic dataset experiments, we use ten random seeds, taken from [10, 20, ... 100].

%Again, removing the citation for scikit learn
%\cite{scikit-learn}

\subsubsection{Hyperparameters for GANs}

We use MLPs of varying depth and width for our Generators and Discriminators.  Due to the sensitive nature of GAN training, we tune the batch size, depth, and number of hidden dimensions, in addition to the following hyperparameters:
\begin{enumerate}[leftmargin=10pt, itemsep=0.1em, topsep=0pt]
    \item $n_{critic}$, which controls the relative number of training steps between the Dicriminator and the Generator
    \item $\lambda$, the weight of the transport cost in the Generator loss function specific to \cite{black2019fliptest}.
\end{enumerate}

\noindent Collectively, these parameters account for the architecture, loss function, and optimization of the GANs, to see which components of the training pipeline are most influential.  As per the original Fliptest paper, we tune the GAN models using the following metrics:

\begin{enumerate}[leftmargin=10pt, itemsep=0.1em, topsep=0pt]
    \item The Kolmogorov–Smirnov (KS) two-sample test~\cite{Hodges2sample} on the marginal distributions for each feature between the real data $\textbf{x}$ and the generated data $G(\textbf{x})$.  Better GAN models will have a smaller KS-statistic (averaged across features), as the real and generated  distributions will be similar.  
    \item The Mean-Squared Error (MSE) of a linear regression model trained to predict each observable feature from the remaining features.  Better GAN models will have lower MSE values (averaged across features), as they will have captured correlation between the features well.
\end{enumerate}

In Figure \ref{fig:gan-metrics}, we examine the correlation of these GAN metrics on the Synthetic-5 experiment, shown in Table 1.  We see that, while a smaller KS statistic does lead to a higher average precision, the MSE metric is at odds precision score.  In this experiment, the most relevant hyperparameter was the weight of the transport cost $\lambda$: generally, we find the best performance is given by very small values (less than $10^{-4}$).  We use this value when testing fair/unfair models on the real datasets.

\subsection{Scope of Work}\label{sec:scope}

In this section, we collect the assumptions and limitations of our approach:

\begin{itemize}[leftmargin=10pt, noitemsep, topsep=0pt]
    \item To assess the impact of modifying the protected variable $C$, we must train an auxiliary model to predict $C$.  This requires the protected variable to be an explicit variable in the training data.  However, because of regulations, it is not always possible to collect this.  In this case, FTA may be a better choice.
    \item We base our test on definitions of independence.  For certain applications, however, this may be the wrong criteria to use.  For instance, in the presence of label bias, it might make sense to use the protected variable as an input for biased mitigation.  In this case, our technique is not applicable.
    \item Our test leverages correlations between the input data $X$ and the protected variable $C$ in order to identify discrimination.  Path-dependent discrimination on the other hand, is out of the scope of this paper.
    \item The features flagged by fAux as surrogates may have a legitimate business reason for being used.  For example, with the adult dataset, on average women work fewer hours per week, but it is fair to use this feature to predict income. fAux will flag features for being correlated with a protected variable, so that a human validator is needed to interpret the final feature rankings. 
    \item Our analysis employs a single first order Taylor expansion around an input $\textbf{x}$.  We thus only consider discrimination that may be revealed by perturbations in a small neighbourhood around the input.  If the target model does not change appreciably over this neighbourhood (say, for points very far from the decision boundary), it may be necessary to consider  successive first order expansions.  This will be a topic for future investigation.
\end{itemize}

\end{document}